\documentclass[10pt,twocolumn,letterpaper]{article}
\usepackage[T1]{fontenc}
\usepackage{times,graphicx,amsmath,amssymb,booktabs,array,multirow,makecell}
\usepackage[table,dvipsnames]{xcolor}
\usepackage[letterpaper,textwidth=6.875in,textheight=8.875in,columnsep=.3125in]{geometry}

\usepackage[numbers,sort&compress]{natbib}
\usepackage[font=small,labelsep=period]{caption}
\usepackage{subcaption,enumitem,stfloats,algorithm,algpseudocode,afterpage,microtype,xspace}
\usepackage[breaklinks,colorlinks,allcolors=linkblue]{hyperref}
\usepackage[capitalize,nameinlink]{cleveref}
\definecolor{linkblue}{rgb}{0.21,0.49,0.74}
\definecolor{oursrow}{gray}{0.94}
\crefname{section}{Sec.}{Secs.}
\Crefname{section}{Section}{Sections}
\crefname{table}{Tab.}{Tabs.}
\Crefname{table}{Table}{Tables}
\setlist[itemize]{noitemsep,leftmargin=*,topsep=0pt}
\newcommand{\eg}{\emph{e.g.}\xspace}

\newcommand{\Lpose}{\mathcal{L}_{\text{pose}}}

\DeclareMathOperator*{\median}{median}
\newif\iflegacy
\legacyfalse
\newcommand{\NumMeanRelAUC}{8.8\%}
\newcommand{\NumMeanRelFone}{3.9\%}
\newcommand{\NumTrainSecMax}{about 2\,min per scene}

\algrenewcommand\algorithmicrequire{\textbf{Require:}}
\algrenewcommand\algorithmicindent{1.2em}
\makeatletter
\renewcommand{\paragraph}[1]{\if@nobreak\else\vspace{.28em plus .08em minus .04em}\fi\noindent\textbf{#1}\ }
\makeatother
\title{Free Geometry: Refining 3D Reconstruction from Stronger Versions of Itself} 

\author{Yuhang Dai \and Xingyi Yang}
\date{}
\hypersetup{pdftitle={Free Geometry: Refining 3D Reconstruction from Stronger Versions of Itself},pdfauthor={Yuhang Dai and Xingyi Yang}}
\newcommand{\mailmark}{\includegraphics[height=.65em]{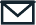}}
\newcommand{\labmark}{\raisebox{.48em}{\includegraphics[height=.66em]{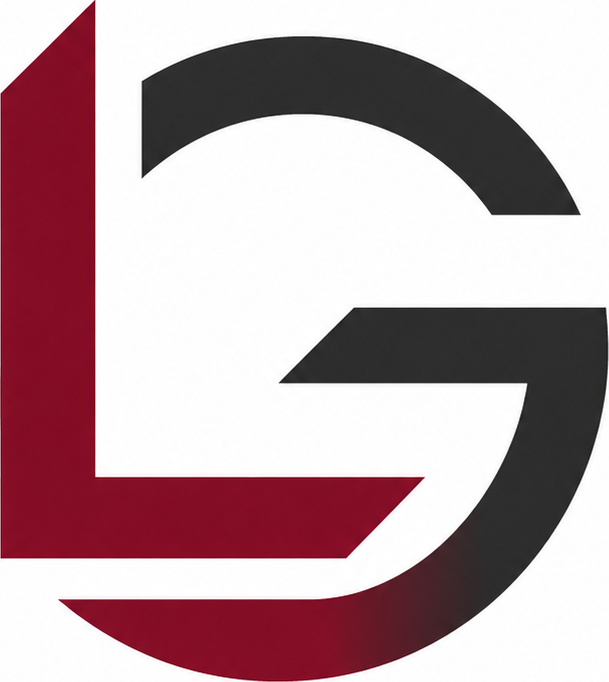}}}
\newcommand{\projectmark}[1]{\includegraphics[height=#1]{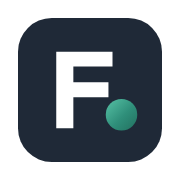}}
\newcommand{\projecturl}{https://free-geometry.github.io/}
\newcommand{\codeurl}{https://github.com/yuhangdai427/Free-Geometry}
\renewenvironment{abstract}{\centerline{\large\bfseries Abstract}\vspace{10pt}\noindent\itshape\ignorespaces}{\par\vspace{10pt}}
\begin{document}
\twocolumn[{%
\begin{center}
\vspace*{.24in}
{\Large\bfseries
\raisebox{-.25em}{\projectmark{1.30em}}\hspace{.35em}Free Geometry: Refining 3D Reconstruction\\[2pt]
from Stronger Versions of Itself\par}
\vspace{22pt}
{\large Yuhang Dai\,\labmark\qquad
Xingyi Yang\,\labmark\,\raisebox{.48em}{\mailmark}\par}
\vspace{3pt}
{\large \href{https://lglabpolyu.github.io/}{\labmark}\,Learning \& Generation Lab, The Hong Kong Polytechnic University\par}
{\normalsize Hung Hom, Kowloon, Hong Kong\par}
\vspace{2pt}
{\small\ttfamily
\href{mailto:yu-hang.dai@connect.polyu.hk}{yu-hang.dai@connect.polyu.hk}\qquad
\href{mailto:xingyi.yang@polyu.edu.hk}{xingyi.yang@polyu.edu.hk}\par}
\vspace{14pt}
{\small
\href{\projecturl}{\raisebox{-.17em}{\projectmark{.95em}}\,Project page}
\enspace$\cdot$\enspace
\href{\codeurl}{\raisebox{-.14em}{\includegraphics[height=.95em]{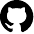}}\,Code}\par}
\vspace{16pt}
  \captionsetup{type=figure}
  \includegraphics[width=\textwidth]{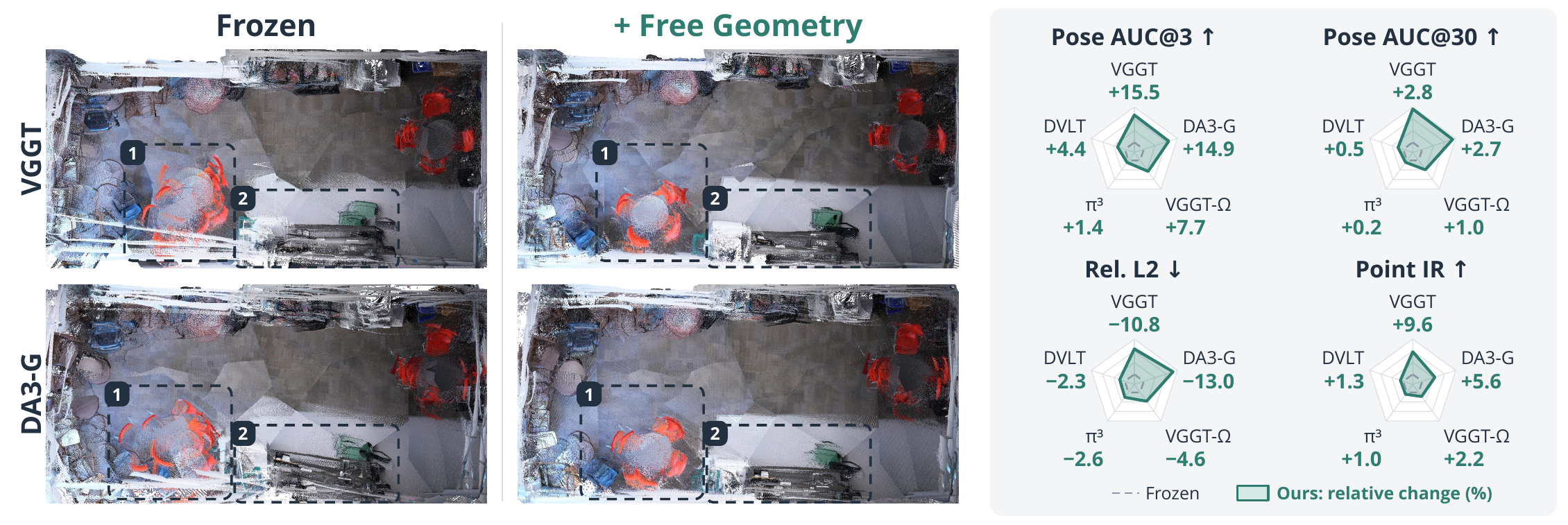} 
  \captionof{figure}{\textbf{Free Geometry} lets feed-forward 3D reconstruction models self-evolve at test time without 3D ground truth. \emph{Left}: VGGT and Depth Anything 3-Giant (DA3-G) on a ScanNet++ scene; boxes mark details improved by Free Geometry. \emph{Right}: relative change on the DVLT benchmark (\cref{tab:dvlt_main,tab:generalize}), one radar per metric with its own linear scale; the dashed ring is the frozen model.}
  \label{fig:teaser}
\end{center}
\vspace{0.5em}
}]

{\renewcommand{\thefootnote}{\mailmark}\footnotetext[0]{Corresponding author.}}
\begin{abstract}

Feed-forward 3D reconstruction models are efficient but rigid: once trained, they run zero-shot and cannot adapt to the test scene.
As a result, plausible-looking reconstructions often contain errors under domain shifts, occlusions, specularities and ambiguous cues.
To address this, we introduce \textbf{Free Geometry}, a framework that enables feed-forward 3D reconstruction models to self-evolve at test time without any 3D ground truth or offline retraining.
Our key insight is that a model can become its own teacher at test time. When given more views, the model produces more geometrically consistent
reconstructions. Aggregating multiple such predictions across different reference frames and view subsets further improves their accuracy and reliability, forming a stronger consensus teacher. Leveraging this property, given a test sequence, we mask different subsets of it to construct multiple partial observations. Free Geometry then adapts the model by enforcing camera pose and depth consistency between these partial-view predictions and the teacher's consensus. By updating only the normalization layers, this self-supervision enables fast scene-specific adaptation in about two minutes on a single GPU.
Our approach improves five state-of-the-art foundation models on four datasets, yielding average relative gains of \NumMeanRelAUC{} in camera pose accuracy (AUC@3) and \NumMeanRelFone{} in point map accuracy (inlier ratio). The project page and code repository are linked on the first page.

\end{abstract}

\section{Introduction}
\label{sec:intro}

Recent advances in feed-forward multi-view 3D reconstruction models, such as Depth Anything 3~\cite{da3} and VGGT~\cite{vggt}, have made it possible to reconstruct 3D scenes in real time. Despite their strong zero-shot performance, these models follow a \emph{train-then-freeze} paradigm: once trained on large-scale datasets, their parameters remain fixed during deployment. Consequently, on novel scenes, reconstructions may look plausible yet contain geometric errors under occlusions, specularities and other ambiguous cues.

A straightforward solution to better generalization is to scale training data. However, collecting large-scale, high-quality 3D ground truth across diverse real-world environments is prohibitively expensive and often impractical. This scarcity of 3D supervision makes (re-)training infeasible.

\begin{figure}[t]
    \centering
    \includegraphics[width=\linewidth]{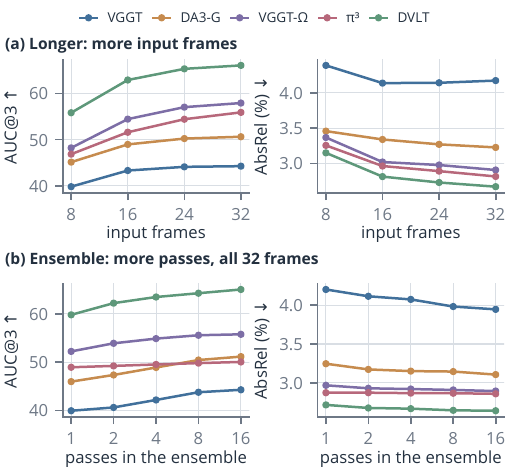}
    \caption{\textbf{A Stronger Teacher of Itself, for Free.} Pose AUC@3 and depth AbsRel of the frozen models, mean over four datasets.}
    \label{fig:intro_compare}
\end{figure}

Our key insight is that a frozen model can build a stronger version of itself from two properties. First, \emph{more views lead to more accurate reconstructions}. As shown in \cref{fig:intro_compare}a, wider viewpoint coverage provides stronger geometric constraints and reduces reconstruction ambiguity. Second, \emph{aggregating predictions leads to more consistent reconstructions}. As shown in \cref{fig:intro_compare}b, even full-sequence predictions vary across frame orders, while aggregating them into a consensus reduces variance and provides more reliable supervision.
Together, the two properties turn the frozen model into a stronger teacher of itself, for free.

Building on this insight, we introduce \textbf{Free Geometry}, a test-time adaptation framework that lets feed-forward models self-evolve on each test scene. The frozen model processes all frames of the scene to produce the teacher, a consensus of cameras and depth over passes with different reference frames and view subsets. The same model with trainable normalization layers~\cite{layernorm} processes masked subsets of the frames to produce the students. Free Geometry optimizes the normalization layers by enforcing camera and depth consistency between the student and teacher predictions of the same frames, a label-free, self-supervised objective.

As a result, Free Geometry adapts at test time in \NumTrainSecMax{} on a single GPU, without labels, and transfers across feed-forward multi-view models. On the four datasets of the DVLT benchmark, it improves five state-of-the-art foundation models, yielding an average relative improvement of \NumMeanRelAUC{} in camera pose accuracy (AUC@3) and \NumMeanRelFone{} in point map accuracy (inlier ratio).

To summarize, the contributions of this work are as follows:
(1) We identify two free sources of a stronger teacher in feed-forward multi-view 3D foundation models: a longer input and a self-ensemble over reference frames and view subsets. (2) We propose \textbf{Free Geometry}, a plug-and-play test-time adaptation framework that supervises masked-view students with this stronger teacher of the model itself, through camera- and depth-level losses on the normalization layers without any 3D ground truth. (3) Experiments on four datasets, five models and three baselines demonstrate that our method delivers low-cost, effective adaptation and improves reconstruction quality and pose accuracy.

\begin{figure*}[t]
    \centering
    \includegraphics[width=\linewidth]{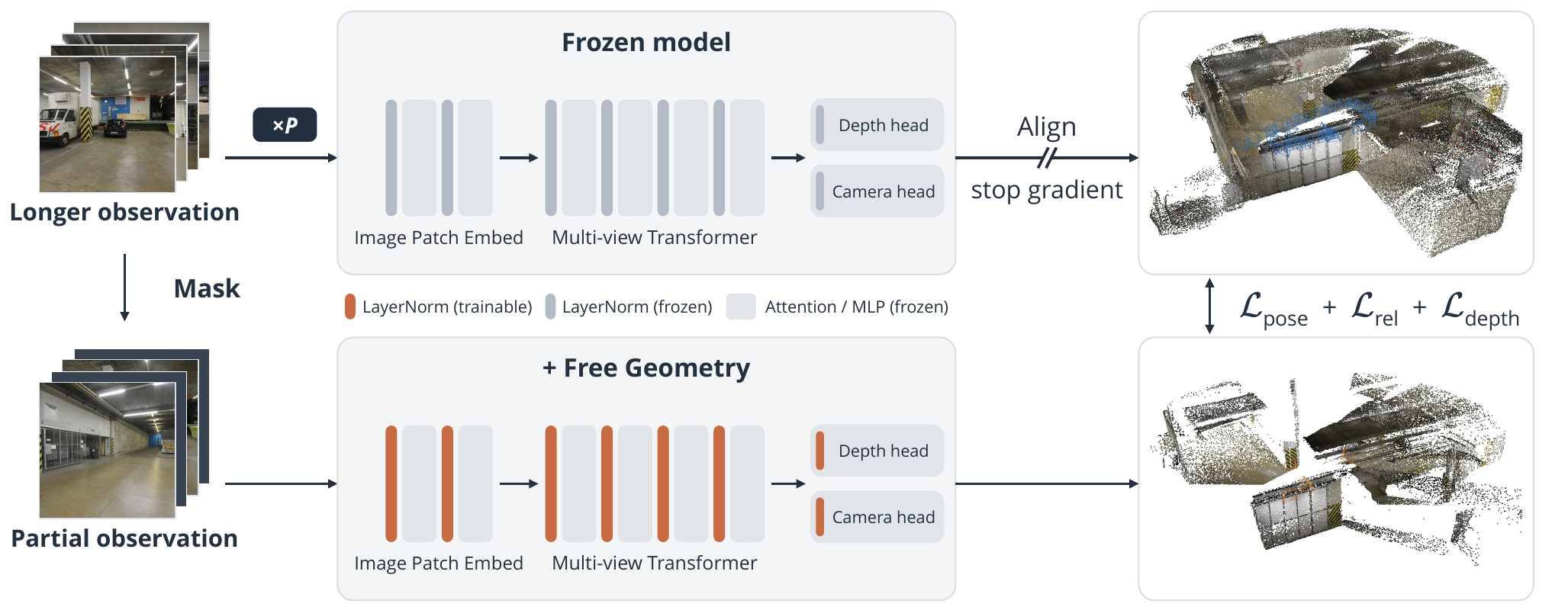}
    \caption{\textbf{Architecture of Free Geometry.} \emph{Top}: all $N$ frames pass $P$ times through the frozen model (different reference orders and near-full subsets); their cameras and depth are aligned, without gradient (double slash), into the consensus (top right). \emph{Bottom}: two partial observations of $M=N/2$ frames (first frame kept) pass through the same model, where only the normalization layers (orange) are trained and all else is frozen (gray). The losses compare their cameras and depth (bottom right) with the consensus.}
    \label{fig:framework}
\end{figure*}

\section{Related Work}
\label{sec:related}

\paragraph{Multi-View Feed-Forward 3D Reconstruction.}
Classical pipelines such as COLMAP~\cite{schoenberger2016sfm,schoenberger2016mvs} and learning-based MVS such as MVSNet~\cite{mvsnet} rely on iterative optimization or cost volumes and need known or estimated camera poses. DUSt3R~\cite{dust3r} regresses point maps from image pairs without poses; VGGT~\cite{vggt} predicts the cameras, depth and point maps of many views with a large geometry transformer; Depth Anything 3~\cite{da3} uses a single plain ViT~\cite{vit} initialized from DINOv2~\cite{dinov2}, with cross-view attention; its largest model, DA3-Giant, has a ViT-g encoder. VGGT-$\Omega$~\cite{vggtomega}, DVLT~\cite{dvlt} and $\pi^3$~\cite{pi3} follow the same paradigm; DVLT predicts ray maps instead of cameras, and $\pi^3$ needs no reference frame, as it is equivariant to the order of the views.

\paragraph{Test-Time Adaptation for 3D Reconstruction.}
Test-time adaptation (TTA) adjusts a pre-trained model using only test data, for example by minimizing prediction entropy through the affine parameters of the normalization layers (TENT~\cite{tent}) or by training auxiliary self-supervised tasks such as rotation prediction~\cite{ttt} and contrastive learning with feature alignment~\cite{tttpp}. In 3D reconstruction, Test3R~\cite{test3r} tunes visual prompts so that two image pairs predict consistent point maps for their shared reference view, TCO~\cite{tco} optimizes the model at test time under external priors such as known camera poses, Self-Geometry~\cite{selfgeo} adapts 3D foundation models toward geometrically consistent predictions with geometric cues from external models, and TTT3R~\cite{ttt3r} updates the memory state of recurrent models for longer sequences. SelfEvo~\cite{selfevo} post-trains offline on unlabeled videos of a target domain with a longer-context prediction as the target. Concurrent work MGE~\cite{mge} improves robustness to occluded views via masked geometric pre-training. Our supervision is a fixed consensus of the model's own passes over all frames of the test scene, a stable target that needs no external model, prior or training data. As in TENT~\cite{tent}, we train only the normalization layers.

\paragraph{Ensembles and Self-Distillation.}
Averaging predictions gives a more reliable target, as in test-time augmentation~\cite{tta_aggregation} and MEMO~\cite{memo}, which minimizes the entropy of the prediction averaged over augmentations; Mean Teacher~\cite{meanteacher} and CoTTA~\cite{cotta} train a student against a weight-averaged teacher. More generally, a student can match the outputs~\cite{hinton_kd}, features~\cite{fitnets} or relations~\cite{rkd} of a stronger model, or the consistency between augmented views under a momentum teacher~\cite{byol,dino,dinov2}. Our teacher is an ensemble of the frozen model's own passes over reference frames and view subsets of the scene, fixed during adaptation.

\section{A Stronger Version of Itself for Free}
\label{sec:method}

We first formulate label-free test-time adaptation for feed-forward 3D reconstruction (\cref{sec:setup}) and then answer three questions: how to build a stronger teacher from the frozen model itself (\cref{sec:teacher}), how to let the student learn from it (\cref{sec:student}), and how to supervise geometry (\cref{sec:losses}).

\subsection{Problem Setup}
\label{sec:setup}
Given a pre-trained feed-forward multi-view reconstruction model $f_\theta$ and a test scene $\{I_1,\dots,I_N\}$, we adapt $f_\theta$ to the scene at test time without ground-truth 3D annotations. For an ordered list of frames, the model predicts the camera-to-world rotation $R_i$, the camera centre $\mathbf{c}_i$, the intrinsics $K_i$ and the depth map $D_i$ of every frame $i$, expressed in the frame of the first input (the reference) and up to one global scale. Each scene is adapted on its own $N$ frames, starting from the original weights, for $T=20$ steps, and the adapted model then predicts the same $N$ frames.

\subsection{How to Find a Stronger Teacher}
\label{sec:teacher}
\begin{table*}[t]
\centering
\footnotesize
\setlength{\tabcolsep}{3pt}
\renewcommand{\arraystretch}{0.95}
\caption{\textbf{Comparison on the DVLT benchmark}: (a) camera pose, \textbf{AUC@3} and \textbf{AUC@30} ($\uparrow$); (b) point map, \textbf{Rel.~L2} ($\downarrow$) and inlier ratio \textbf{IR} (\%, $\uparrow$). Mean over three evaluation seeds (standard deviations in the supplementary material). \textbf{Mean}: relative change (\%) against the frozen model, computed per dataset and averaged over the four datasets with equal weight (for Rel.~L2 a negative change is an improvement); $^\dagger$: statistically better than the best baseline, with the 95\% confidence interval of improvement above zero (stratified bootstrap over scenes within each dataset and over evaluation seeds; supplementary material). Best of each block in \textbf{bold}. $^\star$: reimplemented following the original paper.}
\label{tab:dvlt_main}
{\footnotesize\textbf{(a) Camera pose}}\\[1pt]
\begin{tabular*}{\textwidth}{@{\extracolsep{\fill}}l rr rr rr rr cc}
\toprule
\multirow{2}{*}[-4pt]{\textbf{Method}} & \multicolumn{2}{c}{\textbf{DTU} {\scriptsize(22)}} & \multicolumn{2}{c}{\textbf{ETH3D} {\scriptsize(13)}} & \multicolumn{2}{c}{\textbf{7-Scenes} {\scriptsize(18)}} & \multicolumn{2}{c}{\textbf{ScanNet++} {\scriptsize(50)}} & \multicolumn{2}{c}{\textbf{Mean} {\scriptsize(rel.\,\%)}} \\
\cmidrule(lr){2-3}\cmidrule(lr){4-5}\cmidrule(lr){6-7}\cmidrule(lr){8-9}\cmidrule(lr){10-11}
\multicolumn{1}{r}{\normalfont AUC} & @3$\uparrow$ & @30$\uparrow$ & @3$\uparrow$ & @30$\uparrow$ & @3$\uparrow$ & @30$\uparrow$ & @3$\uparrow$ & @30$\uparrow$ & @3$\uparrow$ & @30$\uparrow$ \\
\midrule
VGGT~\cite{vggt} & 95.6 & 99.8 & 35.3 & 82.4 & 11.5 & 79.7 & 16.0 & 76.3 & -- & -- \\
\quad + Test3R$^\star$~\cite{test3r} & 96.7 & \textbf{99.9} & 36.6 & 82.0 & 11.5 & 79.8 & 15.8 & 75.6 & +0.9 & -0.3 \\
\quad + TCO~\cite{tco} & 88.8 & 99.3 & 30.7 & 81.0 & 11.5 & 79.8 & 17.3 & 76.7 & -2.8 & -0.4 \\
\quad + Self-Geometry$^\star$~\cite{selfgeo} & 78.2 & 97.5 & 22.4 & 71.1 & 7.3 & 69.2 & 10.6 & 68.3 & -31.1 & -9.9 \\
\quad + Ours & \textbf{96.8} & \textbf{99.9} & \textbf{41.3} & \textbf{86.0} & \textbf{12.3} & \textbf{80.6} & \textbf{21.7} & \textbf{80.8} & \textbf{+15.5}$^\dagger$ & \textbf{+2.8}$^\dagger$ \\
\midrule
Depth Anything 3-Giant~\cite{da3} & 85.9 & 99.0 & 58.8 & 81.6 & 13.6 & \textbf{81.3} & 21.9 & 78.1 & -- & -- \\
\quad + Test3R$^\star$~\cite{test3r} & 86.8 & 99.1 & 55.8 & 80.7 & 13.0 & 80.0 & 20.4 & 74.0 & -3.5 & -1.9 \\
\quad + TCO~\cite{tco} & 87.1 & 99.1 & 59.0 & 81.6 & \textbf{13.7} & \textbf{81.3} & 22.9 & 78.5 & +1.9 & +0.2 \\
\quad + Self-Geometry$^\star$~\cite{selfgeo} & 84.8 & 98.9 & 43.8 & 78.9 & 12.2 & 78.0 & 17.5 & 75.1 & -14.1 & -2.8 \\
\quad + Ours & \textbf{91.1} & \textbf{99.6} & \textbf{63.1} & \textbf{84.5} & 13.6 & \textbf{81.3} & \textbf{31.8} & \textbf{83.3} & \textbf{+14.9}$^\dagger$ & \textbf{+2.7}$^\dagger$ \\
\bottomrule
\end{tabular*}
\\[5pt]
{\footnotesize\textbf{(b) Point map}}\\[1pt]
\begin{tabular*}{\textwidth}{@{\extracolsep{\fill}}l rr rr rr rr cc}
\toprule
\multirow{2}{*}[-4pt]{\textbf{Method}} & \multicolumn{2}{c}{\textbf{DTU} {\scriptsize(22)}} & \multicolumn{2}{c}{\textbf{ETH3D} {\scriptsize(13)}} & \multicolumn{2}{c}{\textbf{7-Scenes} {\scriptsize(18)}} & \multicolumn{2}{c}{\textbf{ScanNet++} {\scriptsize(50)}} & \multicolumn{2}{c}{\textbf{Mean} {\scriptsize(rel.\,\%)}} \\
\cmidrule(lr){2-3}\cmidrule(lr){4-5}\cmidrule(lr){6-7}\cmidrule(lr){8-9}\cmidrule(lr){10-11}
 & Rel.~L2$\downarrow$ & IR$\uparrow$ & Rel.~L2$\downarrow$ & IR$\uparrow$ & Rel.~L2$\downarrow$ & IR$\uparrow$ & Rel.~L2$\downarrow$ & IR$\uparrow$ & Rel.~L2$\downarrow$ & IR$\uparrow$ \\
\midrule
VGGT~\cite{vggt} & 0.010 & 95.5 & 0.056 & 57.6 & 0.043 & 70.3 & 0.041 & 62.8 & -- & -- \\
\quad + Test3R$^\star$~\cite{test3r} & 0.010 & 95.4 & 0.055 & 59.3 & 0.041 & 71.1 & 0.043 & 61.7 & -0.0 & +0.6 \\
\quad + TCO~\cite{tco} & \textbf{0.008} & \textbf{97.4} & 0.054 & 52.2 & \textbf{0.037} & \textbf{74.4} & 0.037 & 69.0 & -10.4 & +2.1 \\
\quad + Self-Geometry$^\star$~\cite{selfgeo} & 0.010 & 96.7 & 0.103 & 36.2 & 0.068 & 45.6 & 0.061 & 49.2 & +48.4 & -23.2 \\
\quad + Ours & 0.010 & 95.6 & \textbf{0.046} & \textbf{67.7} & 0.042 & 71.6 & \textbf{0.032} & \textbf{74.6} & \textbf{-10.8} & \textbf{+9.6}$^\dagger$ \\
\midrule
Depth Anything 3-Giant~\cite{da3} & 0.012 & 96.8 & 0.146 & 63.2 & 0.037 & 70.8 & 0.042 & 58.4 & -- & -- \\
\quad + Test3R$^\star$~\cite{test3r} & \textbf{0.010} & \textbf{97.1} & 0.129 & 63.0 & 0.041 & 68.8 & 0.045 & 56.4 & -2.3 & -1.6 \\
\quad + TCO~\cite{tco} & \textbf{0.010} & \textbf{97.1} & 0.144 & 64.5 & \textbf{0.036} & \textbf{73.2} & 0.040 & 59.7 & -7.4 & +1.9 \\
\quad + Self-Geometry$^\star$~\cite{selfgeo} & 0.012 & 96.7 & 0.153 & 52.2 & 0.044 & 66.4 & 0.044 & 56.9 & +5.3 & -6.6 \\
\quad + Ours & 0.011 & \textbf{97.1} & \textbf{0.116} & \textbf{67.1} & \textbf{0.036} & 72.0 & \textbf{0.033} & \textbf{66.8} & \textbf{-13.0}$^\dagger$ & \textbf{+5.6}$^\dagger$ \\
\bottomrule
\end{tabular*}
\end{table*}

Self-distillation needs a target that is more reliable than a single pass of the model. We build it from a longer input and a consensus over passes (\cref{fig:framework}, top).

\paragraph{A Longer Teacher.}
Predictions of the same target frames improve as the model observes more views, since wider coverage adds geometric constraints and resolves local ambiguity. The teacher therefore sees all $N$ frames of the scene, while each student sees only a subset of them (\cref{sec:student}).

\paragraph{An Ensemble Consensus Teacher.}
A single $N$-frame pass, however, depends on the reference frame and the frame order, so different passes disagree on the cameras. We therefore aggregate $P = 16$ passes of the frozen model into a consensus: the benchmark order, seven orders with another frame as the reference, and eight random near-full subsets of $N - 2$ to $N - 8$ frames (\eg, 30 to 24 frames for $N = 32$). Each pass $p$ is aligned to the first pass by a similarity transform estimated on their shared frames with the Umeyama algorithm~\cite{evo}, giving $(\hat{R}_i^p, \hat{\mathbf{c}}_i^p, \hat{D}_i^p)$, and the consensus of frame $i$ over the passes $p \in \mathcal{P}_i$ that contain it is
\begin{equation}
\bar R_i=\Pi_{SO(3)}\big(\sum_{p}\hat R^p_i\big),\:
\bar{\mathbf{c}}_i=\operatorname*{med}_{p}\hat{\mathbf{c}}^p_i,\:
\bar D_i=\operatorname*{med}_{p}\hat D^p_i,
\label{eq:consensus}
\end{equation}
where $\Pi_{SO(3)}$ projects onto $SO(3)$ and passes with outlying rotations are dropped (details in the supplementary material). The consensus is computed once per scene and kept fixed.

\subsection{How to Teach the Student Effectively}
\label{sec:student}

A stronger teacher alone does not ensure that the student learns: the gap between student and teacher must be neither too wide to bridge nor so small that the updates fit noise.

\paragraph{Constructing Moderate Partial-View Students.}
A student that sees too few frames is too far from the full-view teacher, while a student that sees all $N$ frames loses the cross-view completion signal of the longer teacher and gains less in the point map (\cref{sec:ablation}). Each step therefore draws two students of $M = N/2$ frames (\eg, 16 frames), random subsets in the benchmark order that keep the first frame, so that their cameras share the reference of the consensus. Only the normalization layers~\cite{layernorm} of the model are trained.

\paragraph{Skipping Updates Under Negligible Gaps.}
When the frozen model is already self-consistent on a subset, student and teacher differ mainly by noise rather than structural error. A student whose median rotation plus translation-direction error to the teacher is below $\tau = 0.5^\circ$ is skipped, which protects accurate reconstructions and saves computation.

\subsection{How to Supervise Geometry}
\label{sec:losses}

Let $(R_i,\mathbf{c}_i,D_i)$ and $(\bar R_i,\bar{\mathbf{c}}_i,\bar D_i)$ denote the cameras and depth of the partial branch and of the consensus for the $M$ unmasked frames. All losses compare relative quantities (relative poses, normalized distances and log-depth up to a shift), so student and teacher need no alignment.

\begin{figure*}[t]
\centering

\includegraphics[width=\linewidth]{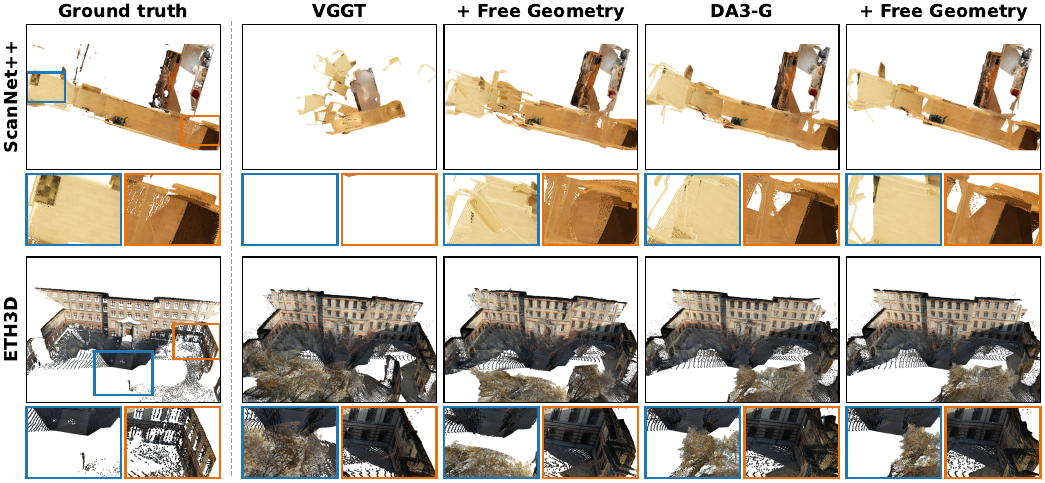}
\caption{\textbf{Qualitative results.} Point clouds of VGGT and DA3-G before and after Free Geometry. Free Geometry recovers more detail and a more correct structure: it unfolds the ScanNet++ room, completes the end of the corridor (orange), and puts the ETH3D side-wing wall (blue) and side wing (orange) in place. The first column is the ground truth.}
\label{fig:pointcloud}
\end{figure*}
\begin{figure*}[t]
\centering

\includegraphics[width=\linewidth]{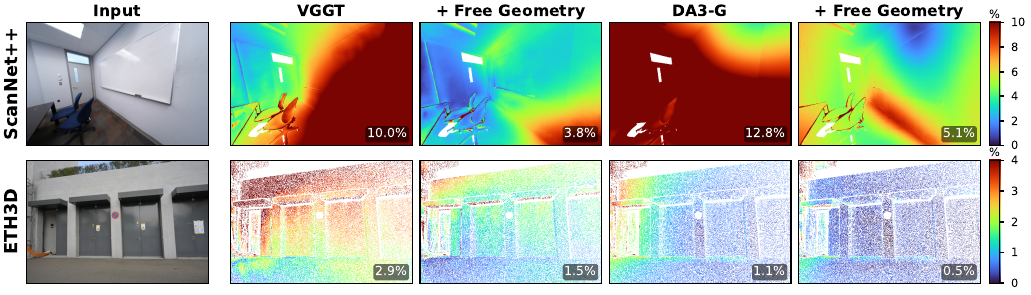}
\caption{\textbf{Per-pixel point-map error} (Rel.~L2) before and after Free Geometry. The number is the frame mean. After adaptation, the error drops over most of the frame for both models, \eg, from 10.0\% to 3.8\% (VGGT) and from 12.8\% to 5.1\% (DA3-G) in the ScanNet++ room.}
\label{fig:depth}
\end{figure*}

\begin{figure*}[t]
\centering
\includegraphics[width=\linewidth]{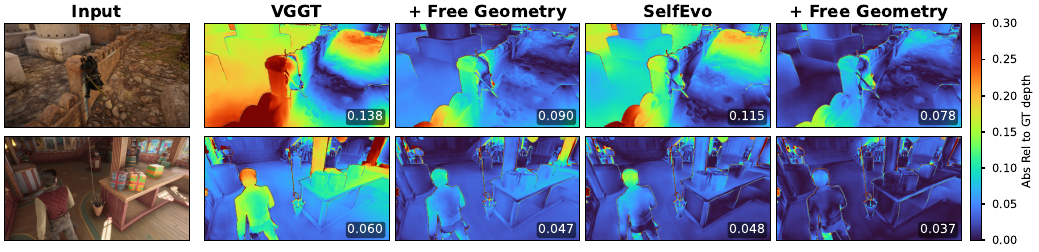}
\caption{\textbf{Out-of-domain scenes} (OmniGeo). Depth error of VGGT and of the domain-adapted SelfEvo, before and after Free Geometry. Without target-domain training data, Free Geometry removes the large errors of VGGT on the foreground structure and on the person, and further lowers the error of SelfEvo over the whole frame. The number is the Abs Rel of the whole sequence.}
\label{fig:qual_ood}
\end{figure*}

\paragraph{Intra-frame Consistency Loss.}
The most direct form of adaptation requires the partial branch's depth to match the consensus depth in log space, up to one shift $\beta$ per input:
\begin{equation}
\mathcal{L}_{\text{depth}}=\operatorname*{mean}_{i,\,p}\Big|\log D_i(p)-\log\bar D_i(p)-\beta\Big|.
\label{eq:depth}
\end{equation}

\paragraph{Cross-frame Relational Pose Loss.}
While $\mathcal{L}_{\text{depth}}$ aligns each frame on its own, it ignores the spatial topology induced by the sequence. With relative rotations $R_{ij}=R_j^{\!\top}R_i$ and relative translation directions $\hat{\mathbf{t}}_{ij}$ over all pairs $i<j$ of the $M$ frames, the relative pose loss is
\begin{equation}
\Lpose = \operatorname*{mean}_{i<j}\ \big\|R_{ij}-\bar R_{ij}\big\|_F + \big|\sin\angle(\hat{\mathbf{t}}_{ij},\hat{\bar{\mathbf{t}}}_{ij})\big|,
\label{eq:pose_a}
\end{equation}
which, like the AUC of the DVLT evaluator ($\min(\theta, 180^\circ-\theta)$ for the translation direction), ignores its sign. A second term $\mathcal{L}_{\text{rel}}$ matches the relations between the camera centres~\cite{rkd}: the pairwise distances $\tilde d_{ij}$ (normalized by their mean), the angles $\phi^k_{ab}$ at camera $k$ between two others, and the camera spread relative to the depth, $s=\log\sigma(\mathbf{c})-\log\operatorname{mean}(D)$:
\begin{equation}
\begin{aligned}
\mathcal{L}_{\text{rel}}=\;&1.5\operatorname*{mean}_{i<j}H\big(\tilde d_{ij}-\tilde{\bar d}_{ij}\big)\\
+\;&1.5\operatorname*{mean}_{k,\,a<b}H\big(\cos\phi^k_{ab}-\cos\bar\phi^k_{ab}\big)+\big(s-\bar s\big)^2,
\end{aligned}
\label{eq:rel}
\end{equation}
with $H$ the Huber function ($\delta=0.2$).

\paragraph{Combining the Terms.}
In every step, each term is averaged over the students that are not skipped (\cref{sec:student}), and its gradient with respect to $\theta$ is computed on its own. The three gradients are normalized to unit length and added, so that no term dominates the update whatever its scale,
\begin{equation}
\mathbf{g} = \sum_{k\in\{\text{depth},\,\text{pose},\,\text{rel}\}}\frac{\nabla_\theta \mathcal{L}_k}{\|\nabla_\theta \mathcal{L}_k\|},
\label{eq:combine}
\end{equation}
and $\mathbf{g}$ is clipped to norm 1 before the AdamW~\cite{adamw} update.

\section{Experiments}
\label{sec:experiments}

\subsection{Experimental Setup}
\label{sec:setup_exp}
\paragraph{Datasets.}
We evaluate Free Geometry on the benchmark of DVLT~\cite{dvlt} (\cref{tab:dvlt_main}), with its released data loaders and evaluation code: \textbf{DTU}~\cite{dtu} (22 object-centric scans), \textbf{ETH3D}~\cite{eth3d} (13 indoor and outdoor scenes with laser-scan ground truth), \textbf{7-Scenes}~\cite{7scenes} (18 indoor RGB-D test sequences) and \textbf{ScanNet++}~\cite{scannetpp} (50 large indoor scenes). For ScanNet++, a training dataset of several of these models, we follow the clean scene-level split of DVLT, whose test scenes are not in the training data~\cite{dvlt}.
Each test scene has $N=32$ frames, drawn by the benchmark's uniform sampler for all methods. (Use all if a test scene has fewer frames.)

\paragraph{Evaluations.} We report AUC@3 and AUC@30, the area under the cumulative curve of the pairwise camera error (the larger of the rotation and the translation-direction error). The predicted depth is unprojected with the predicted cameras into one global point map, which is aligned with ground truth by a single similarity transform to assess the relative $\ell_2$ distance (Rel.~L2) and the inlier ratio at $3\%$ relative error (IR). All rows, including the frozen models, run in fp16.

\paragraph{Baselines.} We compare with the frozen VGGT~\cite{vggt} and Depth Anything 3-Giant (DA3-G)~\cite{da3} and with the test-time adaptation methods TCO~\cite{tco}, Self-Geometry~\cite{selfgeo} and Test3R~\cite{test3r} on the same frames. TCO runs with its official code in its ground-truth-free mode, with the frozen model's predictions as priors. Test3R, whose code covers only DUSt3R, and Self-Geometry, which has no code, are reimplemented for VGGT and DA3-G with careful tuning and debugging; all code is in the supplementary material.

\paragraph{Adaptation Details.}
The consensus uses all benchmark frames (16 passes); each step trains the normalization layers on two students of 16 frames ($\tau=0.5^\circ$), for 20 steps of AdamW~\cite{adamw} in fp16 with a peak learning rate of $3\times10^{-4}$; all models and datasets share this protocol and run on a single NVIDIA RTX PRO 6000 GPU (96\,GB).

\subsection{Scene-Level Adaptation}
\label{sec:scene}

Each scene is adapted on $N$ of its frames and then evaluated on these frames, as in prior test-time adaptation methods such as Test3R~\cite{test3r}; \cref{tab:dvlt_main} reports the comparison.

\paragraph{Quantitative Results.}
Free Geometry statistically improves the pose and the point map of both models on every mean metric, consistently over three evaluation seeds; the 95\% confidence intervals of these gains exclude zero (supplementary material). 
TCO is designed to exploit external priors (\cref{sec:related}); without them, it improves mainly the point map and lowers the AUC@3 of VGGT by 2.8\%. Test3R stays close to the frozen model (AUC@3 +0.9\%), and Self-Geometry lowers the accuracy on most scenes despite our extensive debugging of its reimplementation. Across the four datasets, Free Geometry improves all mean metrics of both models over the frozen baselines. It significantly outperforms the strongest adaptation baseline on all mean metrics except VGGT Rel.~L2 against TCO.

\paragraph{Qualitative Results.} In \cref{fig:pointcloud}, Free Geometry corrects the structural errors of the frozen models and recovers more detail, \eg, it unfolds the ScanNet++ room that frozen VGGT folds into overlapping pieces. Per pixel (\cref{fig:depth}), the error drops over most of each frame.

\begin{table}[t]
\centering
\footnotesize
\setlength{\tabcolsep}{3.4pt}
\caption{\textbf{Out-of-domain data}: VGGT and the domain-adapted SelfEvo~\cite{selfevo} on OmniGeo~\cite{omniworld}, evaluated by pose AUC~$\uparrow$, depth Abs~Rel~$\downarrow$ and $\delta{<}1.25$~$\uparrow$. \emph{Adapts on}: data used for adaptation.}
\label{tab:ood}
\begin{tabular}{l l cc cc}
\toprule
 & & \multicolumn{2}{c}{\textbf{Pose AUC}} & \multicolumn{2}{c}{\textbf{Depth}} \\
\cmidrule(lr){3-4}\cmidrule(lr){5-6}
\textbf{Method} & \textbf{Adapts on} & @5$\uparrow$ & @30$\uparrow$ & Abs Rel$\downarrow$ & $\delta\uparrow$ \\
\midrule
VGGT & -- & 45.2 & 72.7 & 0.180 & 0.760 \\
\quad + Ours & test scene & \textbf{52.7} & \textbf{80.0} & \textbf{0.156} & \textbf{0.811} \\
\midrule
SelfEvo & domain videos & 58.3 & 87.3 & 0.124 & 0.867 \\
\quad + Ours & both & \textbf{64.4} & \textbf{91.6} & \textbf{0.118} & \textbf{0.875} \\
\bottomrule
\end{tabular}
\end{table}

\begin{figure}[t]
\centering
\includegraphics[width=\linewidth]{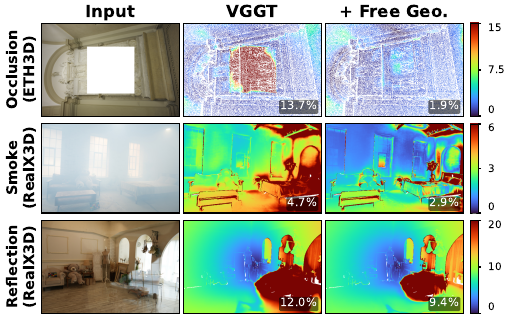}
\caption{\textbf{Challenging scenes} (occlusion, smoke, reflection): input and error of VGGT before and after Free Geometry. Free Geometry repairs the occluded region; under smoke it corrects the walls and furniture that the frozen model misplaces, and under glass reflections it lowers the error on most of the frame.}
\label{fig:qual_hard}
\vspace{0.6em}

{\renewenvironment{table}[1][]{\captionsetup{type=table}}{}\begin{table}[t]
\centering
\footnotesize
\setlength{\tabcolsep}{3.2pt}
\caption{\textbf{Challenging scenes} (occlusion, transparency, real degradations) with VGGT: pose AUC~$\uparrow$, Rel.~L2~$\downarrow$ and inlier ratio (IR, \%)~$\uparrow$ of the global point map. ETH3D with 30\% of the views occluded; B-TOGE: point map on the objects only; RealX3D: mean over its capture conditions (per condition in the supplementary material).}
\label{tab:hard}
\begin{tabular}{l cc cc}
\toprule
 & \multicolumn{2}{c}{\textbf{Pose AUC}} & \multicolumn{2}{c}{\textbf{Point map}} \\
\cmidrule(lr){2-3}\cmidrule(lr){4-5}
\textbf{Method} & @3$\uparrow$ & @30$\uparrow$ & Rel.~L2$\downarrow$ & IR$\uparrow$ \\
\midrule
\multicolumn{5}{l}{\emph{ETH3D + occlusion~\cite{mge}}} \\
\quad VGGT & 34.6 & 81.2 & 0.0750 & 53.5 \\
\quad + Ours & \textbf{38.7} & \textbf{84.7} & \textbf{0.0604} & \textbf{62.4} \\
\midrule
\multicolumn{5}{l}{\emph{Transparent objects (B-TOGE~\cite{weibel2023challenges,btoge})}} \\
\quad VGGT & 65.4 & 96.5 & 0.1157 & 34.8 \\
\quad + Ours & \textbf{78.8} & \textbf{98.3} & \textbf{0.1142} & \textbf{34.9} \\
\midrule
\multicolumn{5}{l}{\emph{RealX3D~\cite{realx3d} (low light, smoke, defocus, etc.)}} \\
\quad VGGT & 84.9 & 98.8 & 0.0308 & 83.9 \\
\quad + Ours & \textbf{85.8} & \textbf{98.9} & \textbf{0.0304} & \textbf{84.3} \\
\bottomrule
\end{tabular}
\end{table}
}
\end{figure}
\subsection{Out-of-Domain and Challenging Scenes}
\label{sec:ood}

\paragraph{Out-of-Domain Datasets.} Free Geometry remains effective on out-of-domain scenes, and further improves a model already adapted to the target domain.
\cref{tab:ood,fig:qual_ood} compare with SelfEvo~\cite{selfevo}, which adapts a model offline to a whole domain of unlabeled videos, whereas Free Geometry adapts to each test scene alone. OmniGeo~\cite{omniworld} consists of rendered video-game environments, a domain far from the real scans the models are trained on; we follow the protocol of~\cite{selfevo}. Free Geometry improves both the original VGGT and the domain-adapted SelfEvo model (\cref{tab:ood}). In some cases, adapting the frozen VGGT yields lower depth error than post-trained SelfEvo (\cref{fig:qual_ood}).

\paragraph{Occlusion, Transparency and Real Degradations.} Free Geometry also assists under challenging conditions.
\cref{tab:hard} tests VGGT where the cues of reconstruction break: ETH3D with 30\% of the views occluded (protocol of MGE~\cite{mge}), transparent objects (B-TOGE~\cite{weibel2023challenges,btoge}; point map on the objects only) and the real degradations of RealX3D~\cite{realx3d} (\eg, smoke, defocus and low light). Free Geometry improves the cameras and the point map on average in all three settings in \cref{tab:hard}. \cref{fig:qual_hard} shows the per-pixel Rel.~L2 (occlusion by a white square as in~\cite{mge}): the error drops in the occluded square, smoke and reflections.

\subsection{Generalization to Other Feed-Forward Models}
\label{sec:generalize}

Free Geometry also transfers to VGGT-$\Omega$~\cite{vggtomega}, $\pi^3$~\cite{pi3} and DVLT~\cite{dvlt} using the same protocol and hyperparameters as VGGT and DA3-G. Two models need small loss or formulation changes because of architectural differences (supplementary material). DVLT predicts a ray map rather than camera parameters by default, so its pose terms use the rotation fitted to the predicted rays, and $\pi^3$ has no reference frame, so its consensus uses only near-full subsets. These changes remain within the unified Free Geometry protocol. Our method yields statistically meaningful gains across all three models on every mean metric (confidence intervals in the supplementary material). Our method also sharpens the details that the frozen models blur or duplicate (\cref{fig:qual_models}): the ScanNet++ walls for $\pi^3$, the ETH3D window frames for VGGT-$\Omega$, and the desk edge in 7-Scenes for DVLT.

\section{Analysis}
\label{sec:analysis}

\subsection{Ablation Study}
\label{sec:ablation}

We ablate each component of Free Geometry on VGGT and ETH3D over three evaluation seeds (\cref{tab:ablation4}); every row is the adapted model. Ablations on DA3-G and on the other datasets are in the supplementary material.

\begin{figure}[t]
\centering
\includegraphics[width=\linewidth]{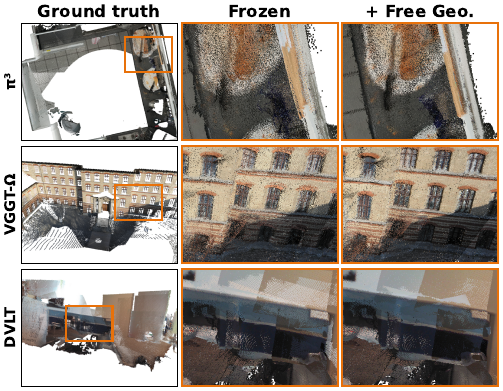}
\caption{\textbf{Point clouds of other models}: $\pi^3$, VGGT-$\Omega$ and DVLT, frozen and with Free Geometry (enlargements of the orange box). Free Geometry sharpens the details of the point maps.}
\label{fig:qual_models}
\vspace{0.8em}
{\renewenvironment{table}[1][]{\captionsetup{type=table}}{}\begin{table}[t]
\centering
\footnotesize
\setlength{\tabcolsep}{4pt}
\caption{\textbf{Generalization to other feed-forward models}: frozen and with Free Geometry, mean over the four datasets and three evaluation seeds, with the protocol and hyperparameters of VGGT and DA3-G; per dataset in the supplementary material. In parentheses: relative change (\%), computed per dataset and averaged over the four datasets, as in the Mean column of \cref{tab:dvlt_main}.}
\label{tab:generalize}
\begin{tabular}{l cc cc}
\toprule
 & \multicolumn{2}{c}{\textbf{Pose AUC}} & \multicolumn{2}{c}{\textbf{Point map}} \\
\cmidrule(lr){2-3}\cmidrule(lr){4-5}
\textbf{Method} & @3$\uparrow$ & @30$\uparrow$ & Rel.~L2$\downarrow$ & IR$\uparrow$ \\
\midrule
VGGT-$\Omega$~\cite{vggtomega} & 52.1 & 91.0 & 0.0245 & 83.8 \\
\quad + Ours & \textbf{55.7}\,{\scriptsize(+7.7)} & \textbf{91.9}\,{\scriptsize(+1.0)} & \textbf{0.0234}\,{\scriptsize($-$4.6)} & \textbf{85.6}\,{\scriptsize(+2.2)} \\
\midrule
$\pi^3$~\cite{pi3} & 49.6 & 90.1 & 0.0225 & 83.9 \\
\quad + Ours & \textbf{50.2}\,{\scriptsize(+1.4)} & \textbf{90.3}\,{\scriptsize(+0.2)} & \textbf{0.0219}\,{\scriptsize($-$2.6)} & \textbf{84.6}\,{\scriptsize(+1.0)} \\
\midrule
DVLT~\cite{dvlt} & 59.8 & 93.4 & 0.0221 & 85.1 \\
\quad + Ours & \textbf{62.8}\,{\scriptsize(+4.4)} & \textbf{93.9}\,{\scriptsize(+0.5)} & \textbf{0.0215}\,{\scriptsize($-$2.3)} & \textbf{86.1}\,{\scriptsize(+1.3)} \\
\bottomrule
\end{tabular}
\end{table}}
\end{figure}

\paragraph{What Makes a Stronger Teacher?}
Blocks~(a) and~(b) evaluate our two progressive teacher sources. Using all 32 frames in a \emph{single} pass~(b) yields virtually no pose gain on average over frozen VGGT because of the noisy pose supervision. The single-pass teacher still improves the point-map inlier ratio over frozen VGGT. Aggregating passes into a consensus stabilizes supervision. Expanding teacher views from 16 to 32 frames progressively lifts AUC@3 to 41.3. Masking student inputs so that the teacher sees more views yields larger point-map gains. Thus, consensus provides essential pose stability while longer observations enrich 3D geometry.

\paragraph{How Can the Student Learn Effectively?}
Block~(c) varies the student length $M$. With sparse views ($M = 4$), the gap to the 32-frame teacher is too wide and student geometry fails to follow the teacher within 20 steps. Conversely, removing masking ($M = 32$) improves pose via pure ensemble distillation but forfeits cross-view completion from a longer teacher, lowering point-map inlier ratio at higher compute per step. Sampling half the views ($M = 16$) best balances learnability with cross-view geometric supervision, showing both teacher sources at work: consensus steadies cameras while fuller context sharpens 3D geometry.

\paragraph{Are All Loss Terms Needed?}
Block~(d) removes one loss at a time. Dropping $\mathcal{L}_{\text{pose}}$ forfeits most of the pose gain, while removing $\mathcal{L}_{\text{rel}}$ lowers the inlier ratio across all seeds. Notably, removing $\mathcal{L}_{\text{depth}}$ degrades both point-map error and camera pose, showing that intra-frame depth and cross-frame pose supervision mutually reinforce shared representations. Therefore, we use all three losses to supervise.

\paragraph{Which Parameters Should Be Adapted?}
Block~(e) and Tab.~\ref{tab:cost3} compare tuning normalization layers against adapting all backbone weights. Full fine-tuning reaches higher accuracy but updates $3{,}900\times$ more parameters and nearly triples peak memory. For efficiency, we update only the normalization layers; full fine-tuning offers further gains at a higher computational cost.

\begin{table}[t]
\centering
\footnotesize
\setlength{\tabcolsep}{3pt}
\caption{\textbf{Ablations} of Free Geometry on VGGT, ETH3D (13 scenes, mean over three evaluation seeds). Every row is the adapted model; each block changes one component of the default: the teacher it learns from (a, b), the student length (c), the losses (d) and the trained parameters (e). Best of each block in \textbf{bold}.}
\label{tab:ablation4}
\resizebox{\linewidth}{!}{%
\begin{tabular}{l cccc}
\toprule
\textbf{Variant} & AUC@3$\uparrow$ & AUC@30$\uparrow$ & Rel.~L2$\downarrow$ & IR$\uparrow$ \\
\midrule
Frozen VGGT & 35.3 & 82.4 & 0.0556 & 57.6 \\
\midrule
\multicolumn{5}{l}{\emph{(a) Teacher views}} \\
\quad 16 frames & 38.7 & 85.3 & 0.0498 & 66.1 \\
\quad 24 frames & 40.4 & 85.7 & 0.0478 & 65.9 \\
\quad all frames (default) & \textbf{41.3} & \textbf{86.0} & \textbf{0.0460} & \textbf{67.7} \\
\midrule
\multicolumn{5}{l}{\emph{(b) Teacher aggregation}} \\
\quad single pass & 35.4 & 82.4 & 0.0573 & 58.2 \\
\quad consensus (default) & \textbf{41.3} & \textbf{86.0} & \textbf{0.0460} & \textbf{67.7} \\
\midrule
\multicolumn{5}{l}{\emph{(c) Student length $M$ (frames)}} \\
\quad $M=4$ & 35.3 & 83.5 & 0.0477 & 61.2 \\
\quad $M=8$ & 37.9 & 84.5 & 0.0507 & 65.8 \\
\quad $M=16$ (default) & 41.3 & 86.0 & \textbf{0.0460} & \textbf{67.7} \\
\quad $M=32$ (no masking) & \textbf{42.5} & \textbf{86.3} & 0.0475 & 65.6 \\
\midrule
\multicolumn{5}{l}{\emph{(d) Loss terms}} \\
\quad w/o $\Lpose$ & 38.2 & 84.5 & 0.0540 & 64.4 \\
\quad w/o $\mathcal{L}_{\text{rel}}$ & 41.3 & 85.8 & 0.0470 & 67.0 \\
\quad w/o $\mathcal{L}_{\text{depth}}$ & 38.7 & 86.0 & 0.0573 & 63.5 \\
\quad all three (default) & \textbf{41.3} & \textbf{86.0} & \textbf{0.0460} & \textbf{67.7} \\
\midrule
\multicolumn{5}{l}{\emph{(e) Trainable parameters}} \\
\quad all backbone weights & \textbf{42.9} & \textbf{86.7} & \textbf{0.0398} & \textbf{70.3} \\
\quad normalization layers (default) & 41.3 & 86.0 & 0.0460 & 67.7 \\
\bottomrule
\end{tabular}}
\end{table}

\begin{table}[t]
\centering\footnotesize\setlength{\tabcolsep}{4pt}\renewcommand{\arraystretch}{0.92}
\caption{\textbf{Cost and efficiency} per scene on the most expensive setting, DA3-G on ScanNet++ (32 frames, one GPU): trainable parameters, peak memory and time per scene (frozen: one forward pass; $^\star$: our reimplementation), median over three scenes.}
\label{tab:cost3}
\begin{tabular}{@{}lrrr@{}}
\toprule
Method & Params$\downarrow$ & Memory$\downarrow$ & Time$\downarrow$ \\
\midrule
Frozen DA3-G~\cite{da3} & -- & 10.0\,GB & 0.8\,s \\
\midrule
\multicolumn{4}{@{}l}{\emph{(a) Test-time adaptation baselines}} \\
\quad Test3R$^\star$~\cite{test3r} & 0.64\,M & 31.4\,GB & 31\,s \\
\quad TCO~\cite{tco} & 2.65\,M & 20.9\,GB & 121\,s \\
\quad Self-Geometry$^\star$~\cite{selfgeo} & 15.7\,M & 16.1\,GB & 143\,s \\
\midrule
\multicolumn{4}{@{}l}{\emph{(b) Ours, trainable part}} \\
\quad all backbone weights & 1,136\,M & 51.3\,GB & 149\,s \\
\quad norm.\ layers (default) & 0.29\,M & 17.4\,GB & 125\,s \\
\bottomrule
\end{tabular}
\end{table}

\begin{table}[t]
\centering\footnotesize\setlength{\tabcolsep}{3.5pt}\renewcommand{\arraystretch}{0.92}
\caption{\textbf{Clues Toward Recursive Self-Improvement}: the frozen and the adapted model, in one pass or as a 16-pass consensus (the frozen consensus is the teacher); mean over the four datasets and three evaluation seeds. + Free Geometry (default) is the adapted model of \cref{tab:dvlt_main}. Best of each block in \textbf{bold}.}
\label{tab:teacher}
\begin{tabular}{@{}l c cccc@{}}
\toprule
 & & \multicolumn{2}{c}{\textbf{Pose AUC}} & \multicolumn{2}{c}{\textbf{Point map}} \\
\cmidrule(lr){3-4}\cmidrule(lr){5-6}
\textbf{Model} & Passes & @3$\uparrow$ & @30$\uparrow$ & Rel.~L2$\downarrow$ & IR$\uparrow$ \\
\midrule
\multicolumn{6}{@{}l}{\emph{VGGT}} \\
\quad one pass & 1 & 39.6 & 84.6 & .0374 & 71.5 \\
\quad 16-pass consensus (teacher) & 16 & 44.3 & 87.1 & .0323 & 76.9 \\
\quad + Free Geometry (default) & 1 & 43.0 & 86.8 & .0324 & 77.4 \\
\quad + Free Geometry, consensus & 16 & \textbf{45.1} & \textbf{87.4} & \textbf{.0306} & \textbf{78.6} \\
\midrule
\multicolumn{6}{@{}l}{\emph{DA3-G}} \\
\quad one pass & 1 & 45.0 & 85.0 & .0593 & 72.3 \\
\quad 16-pass consensus (teacher) & 16 & 51.1 & 87.3 & .0496 & 75.8 \\
\quad + Free Geometry (default) & 1 & 49.9 & 87.2 & .0492 & 75.8 \\
\quad + Free Geometry, consensus & 16 & \textbf{52.9} & \textbf{88.2} & \textbf{.0438} & \textbf{77.3} \\
\bottomrule
\end{tabular}
\end{table}

\subsection{Clues Toward Recursive Self-Improvement}
\label{sec:self_improvement}

\cref{tab:teacher} compares the adapted model with its teacher on the four datasets. First, in one pass, the student closes most of the pose gap to its 16-pass teacher (72\% of the AUC@3 gap for VGGT, 80\% for DA3-G) and matches or even surpasses the teacher's inlier ratio. Second, with the same 16 passes, the consensus of the adapted model outperforms the teacher on every metric. Third, this stronger consensus shows potential to serve as the target of a further round of adaptation, offering encouraging clues toward recursive self-improvement that we leave for future study on longer view streams.

\subsection{Limitations and Failure Cases}
\label{sec:limitations}

Free Geometry requires multiple views of a static scene; single-image and dynamic settings remain for future study. Failure cases arise when the frozen model is already highly consistent on a scene: little structural error is left to correct, and the updates risk fitting noise. Skipping students below $\tau = 0.5^\circ$ (\cref{sec:student}) mitigates this: on VGGT over all 103 scenes, it reduces the scenes more than one AUC@3 point below the frozen model from 9 to 7, with the same mean AUC@3 (43.0) with and without skipping.

\section{Conclusion}
\label{sec:conclusion}

We presented \textbf{Free Geometry}, a test-time adaptation framework that improves feed-forward multi-view 3D reconstruction on each test scene \emph{without} 3D ground truth. A fixed consensus of the frozen model's predictions from longer inputs supervises masked partial observations; only normalization layers are trained. With one protocol across models, Free Geometry improves five foundation models on the four DVLT datasets and remains effective on out-of-domain and visually challenging scenes, with about two minutes of adaptation per scene. We hope this use of multi-view consistency and multi-pass ensembles encourages test-time learning for 3D foundation models.

\section*{LLM Usage Statement}
The authors take full responsibility for all submitted content, regardless of any LLM or AI-tool use. LLMs (OpenAI ChatGPT/GPT, Anthropic Claude, Google Gemini and Moonshot Kimi) were used to edit and refine author-written text, including the related work and citations, and to assist in the design, execution and analysis of experiments; the full manuscript was also given to these providers to obtain reviewer-style feedback. Parts of the code were generated with LLMs and reviewed, edited and approved by the authors. LLMs were not used for figures, for data or for research ideas, hypotheses, claims or conclusions. All reported numbers come from the authors' own runs.

\begingroup
\small
\setlength{\bibsep}{2pt plus 1pt}
\bibliographystyle{ieeenat_fullname}
\bibliography{main}
\endgroup
\clearpage
\twocolumn[{\centering\LARGE\bfseries Supplementary Material\par\vspace{1em}}]
\setcounter{section}{0}
\renewcommand{\thesection}{S\arabic{section}}
\providecommand{\theHsection}{}
\renewcommand{\theHsection}{supp.\arabic{section}}
\setcounter{table}{0}
\renewcommand{\thetable}{S\arabic{table}}
\providecommand{\theHtable}{}
\renewcommand{\theHtable}{supp.\arabic{table}}
\setcounter{figure}{0}
\renewcommand{\thefigure}{S\arabic{figure}}
\providecommand{\theHfigure}{}
\renewcommand{\theHfigure}{supp.\arabic{figure}}
\setcounter{equation}{0}
\renewcommand{\theequation}{S\arabic{equation}}
\providecommand{\theHequation}{}
\renewcommand{\theHequation}{supp.\arabic{equation}}
\setcounter{algorithm}{0}
\renewcommand{\thealgorithm}{S\arabic{algorithm}}
\providecommand{\theHalgorithm}{}
\renewcommand{\theHalgorithm}{supp.\arabic{algorithm}}
\noindent The supplementary material is organized as follows.
\begin{itemize}[leftmargin=1.3em, itemsep=2pt, topsep=3pt, parsep=0pt]
    \item \textbf{\cref{s:method} (Method Formulation \& Algorithm)} derives the gauge- and scale-invariant geometric objectives, outlier-trimmed multi-view consensus fusion, and end-to-end per-scene adaptation pipeline (\cref{alg:freegeo_pipeline}), expanding \cref{sec:method} of the main paper.
    \item \textbf{\cref{s:setup} (Unified Protocol, Benchmarks \& Fair Baselines)} specifies the single shared protocol across all five foundation models (\cref{tab:impl}), the multi-seed benchmark evaluation setup, and the baseline implementations and their tuning protocol (\cref{sec:setup_exp,sec:generalize}).
    \item \textbf{\cref{s:results} (Scene-Level Stability \& Statistical Significance)} reports multi-seed standard deviations over shifted evaluation grids (\cref{tab:dvlt_pose_std,tab:dvlt_pmap_std}) and optimization method seeds (\cref{tab:methodseed}), alongside stratified paired-bootstrap 95\% confidence intervals against frozen models and competing baselines (\cref{tab:ci_supp}), substantiating \cref{sec:scene} and \cref{tab:dvlt_main}.
    \item \textbf{\cref{s:generalize_supp} (Cross-Architecture Generalization \& Confidence Intervals)} reports 95\% confidence intervals (\cref{tab:ci_other}) and per-dataset breakdowns (\cref{tab:generalize_full,tab:generalize_full_pmap}) for VGGT-$\Omega$, $\pi^3$, and DVLT (\cref{sec:generalize}, \cref{tab:generalize}).
    \item \textbf{\cref{s:qual} (Extended Qualitative Analysis Across Five Models)} visualizes per-pixel point-map error reductions across all five architectures (\cref{fig:s_qual_err_vggt,fig:s_qual_err_da3,fig:s_qual_err_other}) and fused 32-frame 3D point clouds for VGGT and DA3-G (\cref{fig:s_qual_pc}), expanding \cref{fig:pointcloud,fig:depth,fig:qual_models}.
    \item \textbf{\cref{s:robust} (Out-of-Domain \& Physically Degraded Scenes)} details zero-shot transfer and domain-adaptation complementarity on OmniGeo (\cref{tab:ood}) alongside challenging scenes and per-condition RealX3D breakdowns (\cref{tab:s_realx3d}, expanding \cref{sec:ood,tab:hard}).
    \item \textbf{\cref{s:ablation_full} (Full Multi-Dataset \& Multi-Model Ablations)} extends the ETH3D ablation of \cref{sec:ablation} (\cref{tab:ablation4}) to all four DVLT datasets on both VGGT (\cref{tab:abl_vggt_perds}) and DA3-G (\cref{tab:abl_da3_perds}), together with per-frame depth analysis (\cref{tab:loss_perds}).
\end{itemize}

\section{Mathematical Formulation and Algorithm}
\label{s:method}

We give the geometric objectives (\cref{s:loss}), the frozen-model consensus (\cref{s:consensus}), and the per-scene adaptation algorithm (\cref{alg:freegeo_pipeline}).

\subsection{Self-Supervised Geometric Losses}
\label{s:loss}
\begin{figure*}[!t]
\centering
\includegraphics[width=0.90\textwidth]{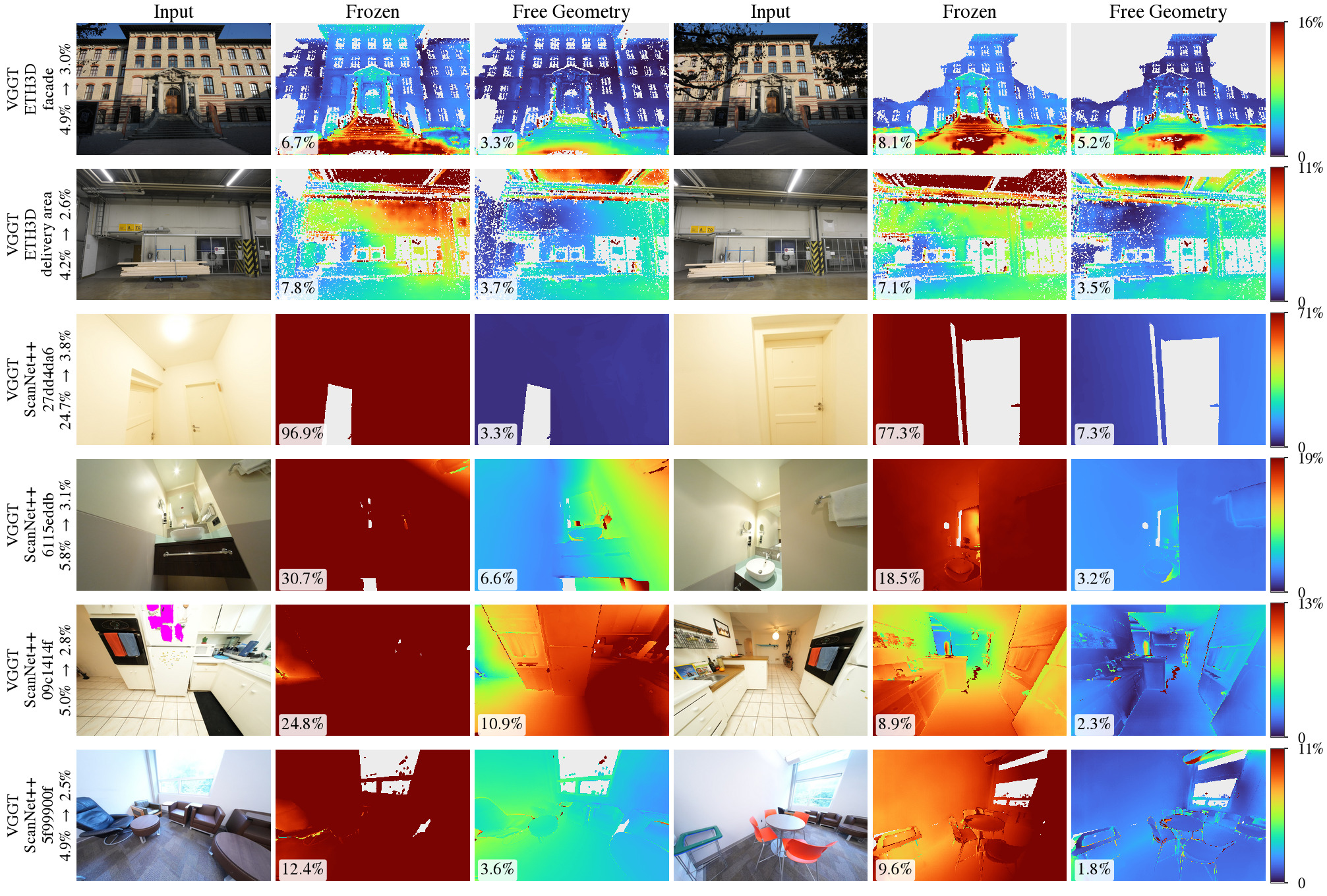}
\caption{\textbf{Per-pixel point-map error of VGGT}, frozen and with Free Geometry, on ETH3D and ScanNet++ (two frames per scene). Each frame shows the input and the error $e(p)$ of the frozen model and of Free Geometry, with its frame mean; the row label gives the scene mean (frozen $\rightarrow$ Free Geometry). The colour range of a row is shared by both models and set to the 95th percentile of the frozen error of that scene. Grey: no ground truth.}
\label{fig:s_qual_err_vggt}
\end{figure*}

Free Geometry adapts a pretrained model $f_\theta$ through a self-supervised geometric objective with three complementary terms. Each term compares the prediction of a partial observation---a masked view subset $v$ of $M$ frames kept in benchmark order and always containing the first frame---against the multi-view consensus of the frozen model (\cref{s:consensus}) restricted to the same frames. The consensus is computed once from the frozen weights $f_{\theta_0}$ and detached; gradients flow exclusively through the affine parameters of the normalization layers of $f_\theta$. As in the main paper, $R_i$ denotes the camera-to-world rotation, $\mathbf{c}_i$ the camera centre, and $D_i$ the predicted depth of frame $i\in v$; bars denote consensus targets. Free Geometry uses mixed precision for model inference and evaluates the geometric losses in \texttt{fp32}.

\paragraph{Intra-frame Consistency Loss.}
Let $\Omega_v=\{(i,p): i\in v,\ 0<\bar D_i(p)<\infty\}$ denote the pixels of the unmasked frames with valid consensus depth, and let $r_i(p)=\log\max(D_i(p),10^{-6})-\log\bar D_i(p)$ denote the log-depth residual. Expanding \cref{eq:depth} of the main paper, the depth consistency objective is
\begin{equation}
\mathcal{L}_{\text{depth}}(v)=\frac{1}{|\Omega_v|}\sum_{(i,p)\in\Omega_v}\big|r_i(p)-\beta_v\big|,
\label{eq:s_depth}
\end{equation}
with a single global shift per partial input given by the mean residual $\beta_v=\frac{1}{|\Omega_v|}\sum_{\Omega_v}r_i(p)$. Removing a single scalar shift leaves one global scale per partial observation free, while tightly constraining both the relative depth scales across frames and the per-pixel geometric shape of every depth map.

\paragraph{Cross-frame Relational Pose Loss.}
For every frame pair $(i,j)\in\mathcal{P}$ ($i<j$) within the $M$ unmasked frames, let $R_{ij}=R_j^\top R_i$ be the relative rotation and $\hat{\mathbf{t}}_{ij}=\mathbf{t}_{ij}/\|\mathbf{t}_{ij}\|$ the unit relative translation direction, where $\mathbf{t}_{ij}=R_j^\top(\mathbf{c}_i-\mathbf{c}_j)$. Expanding \cref{eq:pose_a} of the main paper, the pairwise relative pose objective is
\begin{equation}
\begin{split}
\Lpose(v)=\frac{1}{|\mathcal{P}|}\sum_{(i,j)\in\mathcal{P}}&\Big[\big\|R_{ij}-\bar R_{ij}\big\|_F\\
&+\sqrt{1-(\hat{\mathbf{t}}_{ij}^\top\hat{\bar{\mathbf{t}}}_{ij})^2+\epsilon}-\epsilon'\Big],
\end{split}
\label{eq:s_pose_a}
\end{equation}
where the second term is a smoothed $|\sin|$ of the angle between translation directions ($\epsilon=10^{-6}$, $\epsilon'=10^{-3}$). Squaring the dot product makes the translation penalty invariant to the sign of the translation direction.

To further constrain the global spatial topology of the camera trajectory and bind camera scale to dense depth scale, the relational term (\cref{eq:rel} of the main paper) matches normalized camera-centre relations~\cite{rkd} and camera-to-depth coupling:
\begin{align}
\mathcal{L}_{\text{rel}}(v)&=1.5\,\big(\mathcal{L}_{\text{dist}}+\mathcal{L}_{\text{angle}}\big)+\mathcal{L}_{\text{couple}},\label{eq:s_pose_b}\\
\mathcal{L}_{\text{dist}}&=\frac{1}{|\mathcal{P}|}\sum_{\mathcal{P}}H_{0.2}\Big(\frac{d_{ij}}{\mu}-\frac{\bar d_{ij}}{\bar\mu}\Big),\nonumber\\
\mathcal{L}_{\text{angle}}&=\operatorname*{mean}_{k;\,i<j;\,i,j\neq k}H_{0.2}\big(\cos\angle_k(i,j)-\cos\bar\angle_k(i,j)\big),\nonumber\\
\mathcal{L}_{\text{couple}}&=\Big(\log\frac{\sigma}{m}-\log\frac{\bar\sigma}{\bar m}\Big)^2,\nonumber
\end{align}
where $d_{ij}=\|\mathbf{c}_i-\mathbf{c}_j\|$, $\mu$ is the mean pairwise distance, $\angle_k(i,j)$ is the angle at camera $k$ subtended by cameras $i$ and $j$, and $H_{\delta}(x)=x^2/2$ for $|x|\le\delta$ and $\delta(|x|-\delta/2)$ otherwise ($\delta=0.2$). For a two-view input the angle term is omitted. In $\mathcal{L}_{\text{couple}}$, $\sigma=\big(\frac1M\sum_i\|\mathbf{c}_i-\frac1M\sum_j\mathbf{c}_j\|^2\big)^{1/2}$ is the RMS spread of camera centres and $m$ is the mean of the finite positive predicted depths on $\Omega_v$; $\bar m$ uses the valid consensus depths. These relative quantities allow the student to be compared with its teacher without fitting an alignment during backpropagation. For ray-map architectures such as DVLT~\cite{dvlt}, the pose terms operate on differentiable rotations fitted to the predicted rays (\cref{s:setup}).

\paragraph{Selective Skipping and Gradient Balancing.}
At each adaptation step, we sample $B=2$ partial inputs of $M=\min(16,N)$ frames in benchmark order, each retaining the first frame. To protect already well-reconstructed scenes from overfitting minor noise, any partial input whose median pairwise rotation error plus median translation-direction error to the consensus is below $\tau=0.5^\circ$ is skipped. Each sampled input contributes with weight $1/B$ and skipped inputs contribute zero; if all inputs are skipped, the step performs no parameter update. We use $\varepsilon_g=10^{-30}$ in the gradient denominator. The gradient of each surviving term is normalized to unit $\ell_2$ norm across all trainable parameters before summation (\cref{eq:combine} of the main paper), preventing scale imbalance among rotation, translation, and log-depth terms; the summed update vector is clipped to unit norm.

\paragraph{Optimizer and Learning-Rate Schedule.}
Adapting only the affine weights and biases of normalization layers---spanning the image patch encoder, multi-view transformer, and geometry prediction heads ($358{,}400$ parameters for VGGT; \cref{tab:impl})---regularizes optimization against overfitting a single test scene while keeping memory overhead minimal. We optimize with AdamW~\cite{adamw} ($\beta=(0.9,0.999)$, weight decay $0.05$) for $T=20$ steps with peak learning rate $\eta=3\times10^{-4}$ across all models. Using $w=\max(1,\mathrm{round}(0.2T))=4$ linear warm-up steps, the step-$t$ learning rate follows a cosine decay to $\kappa=10^{-3}$ of the peak:
\begin{equation}
\eta_t=\begin{cases}\eta\,t/w & t\le w,\\[2pt]
\eta\big[\kappa+\tfrac{1-\kappa}{2}\big(1+\cos\tfrac{\pi(t-w)}{T-w}\big)\big] & t>w.\end{cases}
\label{eq:s_lr}
\end{equation}
After final inference on the adapted scene, the optimized normalization parameters are restored. Each benchmark scene starts from the released checkpoint in a fresh process.

\subsection{Multi-View Consensus of the Frozen Model}
\label{s:consensus}
\begin{figure*}[!t]
\centering
\includegraphics[width=0.90\textwidth]{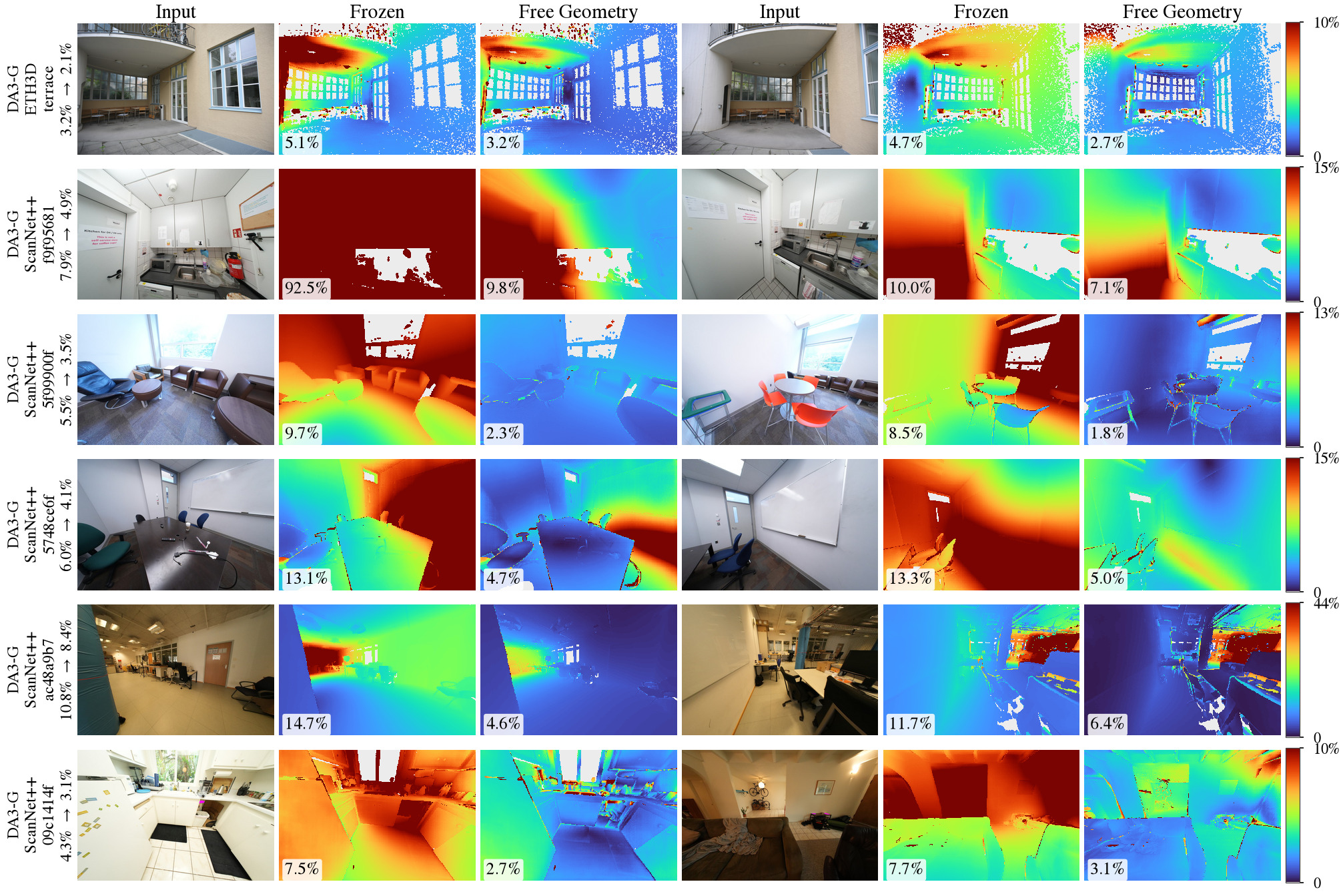}
\caption{\textbf{Per-pixel point-map error of DA3-G}, frozen and with Free Geometry; layout as in \cref{fig:s_qual_err_vggt}.}
\label{fig:s_qual_err_da3}
\end{figure*}

The consensus teacher turns the frozen model $f_{\theta_0}$ into a stronger, lower-variance supervisor once per scene prior to adaptation without ever accessing ground truth. Below we detail the exact member selection, similarity alignment, and outlier-trimmed fusion summarized in \cref{eq:consensus} of the main paper, where $(R^a_i,\mathbf{c}^a_i,K^a_i,D^a_i)$ denotes the prediction of ensemble member $a$ for frame $i$, and member $0$ denotes the baseline frozen forward pass in benchmark order.

\paragraph{Ensemble Members.}
To capture both full-context geometric constraints and diversity across reference views, at $N=32$ the consensus aggregates $|\mathcal{A}|=P=16$ frozen forward passes of the test scene:
\begin{equation}
\mathcal{A}=\{\pi^0\}\;\cup\;\{\pi^r\}_{r=1}^{7}\;\cup\;\{\rho^k\}_{k=1}^{8}.
\label{eq:s_members}
\end{equation}
Member $0$, $\pi^0=(1,\dots,N)$, runs on all $N$ frames (default $N=32$) in benchmark order. The seven reference-shifted passes $\pi^r$ ($r=1,\dots,7$) promote frame $1+\mathrm{round}(rN/8)$ to the front while preserving the relative order of the remaining frames. The eight subset passes $\rho^k$ sample random near-full subsets of 30, 28, 26, and 24 frames (two of each length, capped at $N$) in a fixed-seed random order, diversifying both view context and reference frame. Duplicate reference indices are omitted. At $N=32$, this comprises $16$ passes: $8$ full-sequence passes and $8$ near-full passes ($24$ to $30$ frames at $N=32$). For inherently permutation-equivariant models without a distinguished reference frame such as $\pi^3$~\cite{pi3}, reference-shifted full orders produce identical predictions to $\pi^0$; accordingly, within the exact same $16$-pass budget, its $15$ additional members are drawn as random near-full subsets with the same size cap (one of 30, two of 28, three of 26, and nine of 24 frames).

\paragraph{Similarity Alignment with Hybrid Scale Fallback.}
This section corrects the description of teacher alignment as a point-set Umeyama fit~\cite{evo} in \cref{sec:teacher} of the main paper. The implementation (\texttt{consensus.py}) estimates rotation from camera orientations on shared frames and fits scale and translation to the corresponding camera centres, with the depth-based scale fallback defined below. The benchmark-order pass defines the coordinate frame and has unit scale. For each other member $a$ and the benchmark-order pass $0$, the rotation and centre-based scale are
\begin{equation}
\begin{gathered}
Q_a=\Pi_{SO(3)}\Big(\sum_{i\in a}R^0_i\,{R^a_i}^{\!\top}\Big),\\
s^{c}_a=\frac{\sum_{i\in a}\big\langle \mathbf{c}^0_i-\boldsymbol{\mu}^0_a,\;Q_a(\mathbf{c}^a_i-\boldsymbol{\mu}^a_a)\big\rangle}{\max\!\left(10^{-12},\sum_{i\in a}\|\mathbf{c}^a_i-\boldsymbol{\mu}^a_a\|^2\right)},
\end{gathered}
\label{eq:s_align}
\end{equation}
where $\Pi_{SO(3)}$ projects onto the nearest rotation matrix via SVD (enforcing $\det Q_a=+1$) and $\boldsymbol{\mu}^0_a,\boldsymbol{\mu}^a_a$ are the mean camera centres over shared frames. To remain well-conditioned across arbitrary camera baselines, we validate $s^c_a$ against the dense depth-based scale $s^d_a=\median_{i\in a}\median_{p}\big(D^0_i(p)/D^a_i(p)\big)$, using shared frames with more than 100 pixels where both depths are finite and positive. The inner median uses only those pixels. The \emph{hybrid scale rule} selects
\begin{equation}
s_a=\begin{cases} s^c_a & \text{if } s^c_a>0\ \text{and}\ \big|\log\tfrac{s^c_a}{s^d_a}\big|\le\log 1.5,\\[2pt] s^d_a & \text{otherwise,}\end{cases}
\label{eq:s_scale}
\end{equation}
when a depth estimate is available. If no shared frame supplies a valid depth estimate, the implementation uses the centre-based scale if it is finite and positive; otherwise it reports an undefined alignment. Setting $\mathbf{u}_a=\boldsymbol{\mu}^0_a-s_aQ_a\boldsymbol{\mu}^a_a$ yields aligned predictions $\tilde R^a_i=Q_aR^a_i$, $\tilde{\mathbf{c}}^a_i=s_aQ_a\mathbf{c}^a_i+\mathbf{u}_a$, $\tilde D^a_i=s_aD^a_i$, and $\tilde K^a_i=K^a_i$.

\paragraph{Outlier-Trimmed Consensus Fusion.}
For each frame $i$, let $\mathcal{A}_i$ be the set of members containing frame $i$. We first compute an initial chordal rotation mean $\hat R_i$ and retain only the inlier subset $\mathcal{A}'_i\subseteq\mathcal{A}_i$ whose geodesic angular distance to $\hat R_i$ is at most $\max(2^\circ,\,3\times\text{median angle})$. Fusing over $\mathcal{A}'_i$ via a refined chordal mean for rotation and coordinate-/pixel-wise medians for centres, intrinsics, and depth gives
\begin{equation}
\begin{gathered}
\hat R_i=\Pi_{SO(3)}\Big(\sum_{a\in\mathcal{A}_i}\tilde R^a_i\Big),\quad
\bar R_i=\Pi_{SO(3)}\Big(\sum_{a\in\mathcal{A}'_i}\tilde R^a_i\Big),\\
\bar{\mathbf{c}}_i=\operatorname*{med}_{a\in\mathcal{A}'_i}\tilde{\mathbf{c}}^a_i,\quad
\bar K_i=\operatorname*{med}_{a\in\mathcal{A}'_i}\tilde K^a_i,\quad
\bar D_i(p)=\operatorname*{med}_{a\in\mathcal{A}'_i}\tilde D^a_i(p).
\end{gathered}
\label{eq:s_fuse}
\end{equation}
This two-stage trimming suppresses rotational outliers and produces a clean, geometrically coherent teacher in the native coordinate frame of the frozen model.

\subsection{End-to-End Adaptation Pipeline}
\label{s:pipeline}

\begin{algorithm}[t]
\caption{Free Geometry test-time adaptation}
\label{alg:freegeo_pipeline}
\footnotesize
\begin{algorithmic}[1]
\Require frozen model $f_{\theta_0}$, scene frames $\mathcal{I}=(I_1,\dots,I_N)$, steps $T=20$, peak learning rate $\eta=3\times10^{-4}$, $B=2$ partial inputs per step of $M=\min(16,N)$ frames, threshold $\tau=0.5^\circ$
\State Build consensus $(\bar R_i,\bar{\mathbf{c}}_i,\bar K_i,\bar D_i)_{i=1}^N$ from the frozen passes (\cref{s:consensus})
\State $\theta\leftarrow\theta_0$; unfreeze normalization layer weights and biases only
\For{$t=1,\dots,T$}
    \State Sample $B$ partial inputs $v_1,\dots,v_B$ of $M$ frames in benchmark order keeping frame 1; skip any $v_b$ with pose gap to consensus below $\tau$
    \If{at least one $v_b$ survives}
        \State Accumulate $\mathcal{L}_k$ over retained inputs, with weight $1/B$ per input (\cref{eq:s_depth,eq:s_pose_a,eq:s_pose_b}). Use $(k,w_k)$ as defined in \cref{s:protocol} for DVLT; otherwise use depth, pose and relational terms with $w_k=1$.
        \State $\mathbf{g}\leftarrow\mathrm{clip}_1\!\big(\sum_k w_k\nabla_\theta\mathcal{L}_k/(\|\nabla_\theta\mathcal{L}_k\|+\varepsilon_g)\big)$
        \State Update $\theta$ via AdamW with $\mathbf{g}$ and learning rate $\eta_t$ (\cref{eq:s_lr})
    \EndIf
\EndFor
\State Store $\hat y=f_\theta(\mathcal{I})$ on all $N$ frames in benchmark order; restore the learned normalization parameters to their initial values
\State \Return $\hat y$
\end{algorithmic}
\end{algorithm}

\cref{alg:freegeo_pipeline} summarizes the per-scene workflow; the decoder-specific pose gradient for DVLT is given in \cref{s:protocol}. For every test scene, the frozen model $f_{\theta_0}$ first constructs a detached multi-view consensus from the frozen passes on the selected scene frames. Next, $f_\theta$ adapts solely its normalization layers for $20$ steps by aligning $M$-frame partial observations to this stronger consensus through the balanced three-term objective. Finally, the adapted model runs a single forward pass on all $N$ frames in benchmark order to produce the evaluated cameras and dense 3D reconstruction, after which the normalization parameters are restored.

\section{Protocol, Benchmarks, and Baselines}
\label{s:setup}

We describe the shared adaptation settings, model interfaces, evaluation metrics and baseline configurations used for the comparisons in \cref{tab:dvlt_main,tab:generalize}.

\subsection{Unified Protocol and Model Interfaces}
\label{s:protocol}

\begin{table}[t]
\centering
\footnotesize
\setlength{\tabcolsep}{3pt}
\caption{\textbf{Unified implementation protocol} across all five feed-forward models: trainable normalization parameters, steps, peak learning rate, 16-pass consensus composition, and native camera interface. Every model uses the identical optimizer, cosine schedule, step budget, and partial observations ($B=2$ subsets of up to 16 frames in benchmark order with the first frame fixed, skipped below $\tau=0.5^\circ$).}
\label{tab:impl}
\resizebox{\linewidth}{!}{%
\begin{tabular}{l r c c l l}
\toprule
Model & Trainable & Steps & lr & Consensus & Camera interface \\
\midrule
VGGT & 358{,}400 & 20 & 0.0003 & 1 + 7 orders + 8 subsets & Extrinsics \\
VGGT-$\Omega$ & 354{,}304 & 20 & 0.0003 & 1 + 7 orders + 8 subsets & Extrinsics \\
DA3-G & 292{,}672 & 20 & 0.0003 & 1 + 7 orders + 8 subsets & Extrinsics \\
DVLT & 58{,}368 & 20 & 0.0003 & 1 + 7 orders + 8 subsets & Ray map (Kabsch) \\
$\pi^3$ & 318{,}464 & 20 & 0.0003 & 1 + 15 subsets & Extrinsics \\
\bottomrule
\end{tabular}}
\end{table}

We implement Free Geometry in PyTorch and run all adaptation and evaluation experiments on a single NVIDIA RTX PRO 6000 GPU (96\,GB). During Free Geometry adaptation, only the affine weights and biases of the normalization layers are optimized---covering the image encoder, cross-view transformer, and prediction heads---while every other weight remains frozen (\cref{tab:impl}).

\paragraph{One Protocol Across Five Models.}
On the DVLT benchmark, adaptation starts from the released weights and uses the selected evaluation frames without ground truth. After $T=20$ steps, the model predicts those same frames. As summarized in \cref{tab:impl}, all five models---VGGT~\cite{vggt}, VGGT-$\Omega$~\cite{vggtomega}, DA3-G~\cite{da3}, DVLT~\cite{dvlt}, and $\pi^3$~\cite{pi3}---share the following optimizer and sampling settings: AdamW~\cite{adamw} ($\beta=(0.9,0.999)$, weight decay $0.05$), peak learning rate $\eta=3\times10^{-4}$, four warm-up steps ($\mathrm{round}(0.2T)$) with cosine decay to $10^{-3}\eta$, unit-norm gradient clipping, \texttt{fp16} autocast, and $B=2$ partial observations per step (kept in benchmark order with the first frame fixed, skipped below $\tau=0.5^\circ$). For the default comparison, models use the same per-step seed to sample student views. \Cref{s:methodseed} reports sensitivity to the student-mask seed.

\paragraph{Native Architectural Interfaces Within the Unified Protocol.}
Table~\ref{tab:impl} lists the shared optimization and sampling settings. The following model interfaces determine how cameras are obtained and how consensus members are formed.
\begin{itemize}[leftmargin=1.2em, itemsep=1.5pt, topsep=2pt, parsep=0pt]
    \item \emph{DA3-G~\cite{da3}.} We use the camera decoder and set \texttt{ref\_view\_strategy} to \texttt{first}. The first view of each input is its reference, including after a consensus member reorders the views.
    \item \emph{DVLT~\cite{dvlt}.} DVLT decodes a dense ray map. We take the camera centre from the mean ray origin and fit rotation to the ray directions with a differentiable Kabsch step. The pose loss follows this decoder representation: we normalize the rotation and translation-direction gradients separately and weight each by three; the relational and depth gradients each have unit weight. This decoder-specific pose supervision uses the shared consensus, student sampling, normalization-layer updates, optimizer, and learning-rate schedule of the unified protocol.
    \item \emph{$\pi^3$~\cite{pi3}.} Because $\pi^3$ is inherently permutation-equivariant with no distinguished reference frame, reordering all $N$ views is mathematically invariant. Within the exact same 16-pass consensus budget, its $15$ additional passes are therefore drawn as random subsets as specified in \cref{s:consensus}, while all optimization and masking settings remain unchanged.
\end{itemize}

\subsection{Benchmarks, Seed Sampling, and Metrics}
\label{s:benchmark}
\begin{figure*}[!t]
\centering
\includegraphics[width=0.90\textwidth]{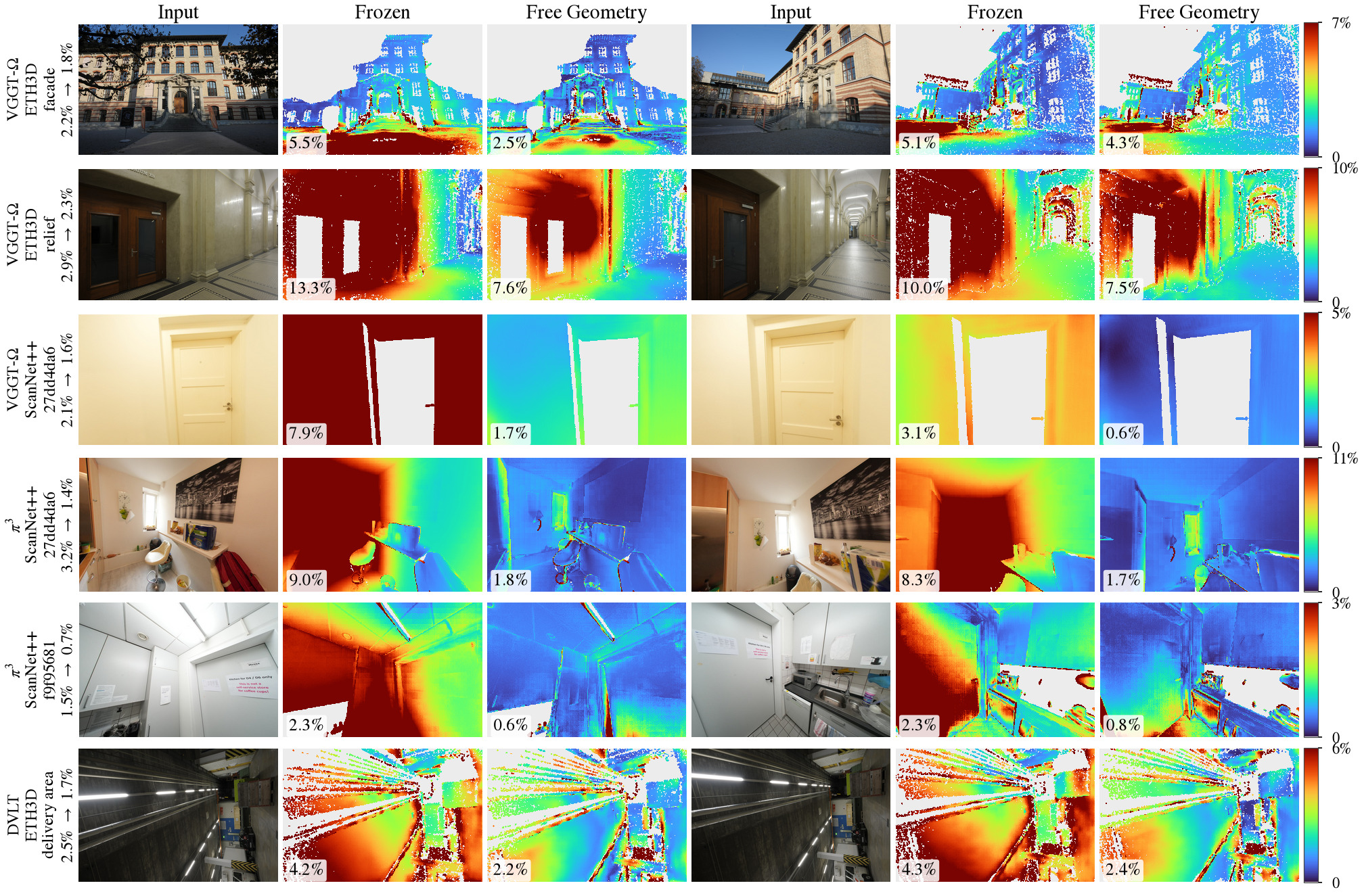}
\caption{\textbf{Per-pixel point-map error of VGGT-$\Omega$, $\pi^3$ and DVLT}, frozen and with Free Geometry; layout as in \cref{fig:s_qual_err_vggt}.}
\label{fig:s_qual_err_other}
\end{figure*}

\paragraph{DVLT Benchmark.}
Our primary evaluation (\cref{tab:dvlt_main,tab:generalize}) uses the benchmark of DVLT~\cite{dvlt} with its official data loaders and evaluators across four diverse datasets: \textbf{DTU}~\cite{dtu} (22 object-centric scans with structured-light ground truth, views ranked from the middle view), \textbf{ETH3D}~\cite{eth3d} (13 indoor/outdoor scenes with undistorted images and laser-scan ground truth), \textbf{7-Scenes}~\cite{7scenes} (18 indoor RGB-D test sequences, Kinect depth registered to the colour camera and filtered to $[0.1,10]\,\text{m}$ following $\pi^3$~\cite{pi3}), and \textbf{ScanNet++}~\cite{scannetpp} (50 large indoor scenes from the clean \texttt{nvs\_sem\_val} split, undistorted images with laser-mesh depth, strictly disjoint from model training splits~\cite{dvlt}). The benchmark sampler selects up to 32 uniformly spaced frames per sequence and uses all available frames for shorter sequences.

\paragraph{Multi-Seed Frame Sampling.}
On the DVLT benchmark, Seed~0 uses the default uniform grid. When subsampling is required, Seeds~1 and~2 shift that grid by one-third and two-thirds of the sampling stride, respectively. All methods use the same frames for each seed.

\paragraph{Evaluation Metrics.}
Standardized, gauge-free evaluation metrics assess both camera trajectory accuracy and global dense 3D structure. For \emph{camera pose}, we compute for every frame pair the angular error of relative rotation and relative translation direction against ground truth (ignoring translation sign ambiguity), take their maximum, and report the Area Under the Cumulative error curve up to $3^\circ$ (\textbf{AUC@3}) and $30^\circ$ (\textbf{AUC@30}). For \emph{3D point maps}, predicted depths are unprojected using predicted cameras into a single global point cloud, which is aligned to ground truth with the benchmark evaluator's point-set Umeyama fit. For every valid point $p$, the relative error is $e(p)=\|\hat{\mathbf{x}}_p-\mathbf{x}_p\|/\|\mathbf{x}_p\|$; \textbf{Rel.~L2} is the mean of $e(p)$ and the inlier ratio \textbf{IR} is the percentage of points with $e(p)<3\%$, averaged over scenes per dataset. In the main paper, the Mean columns of \cref{tab:dvlt_main} and the parenthesized values in \cref{tab:generalize} report relative changes against frozen, computed per dataset and evaluation seed and averaged equally over datasets and seeds.

\paragraph{Additional Evaluations.}
We also evaluate occluded ETH3D~\cite{mge}, transparent objects in B-TOGE~\cite{weibel2023challenges,btoge}, and physical image degradations in RealX3D~\cite{realx3d} (\cref{tab:hard}). On B-TOGE, pose is evaluated over the full image and point maps over transparent objects. Results for OmniGeo and the RealX3D conditions appear in \cref{s:robust}.

\subsection{Fair Implementation and Tuning of Baselines}
\label{s:baselines}
\label{s:base_impl}
\begin{table*}[!t]
\centering
\footnotesize
\setlength{\tabcolsep}{2.6pt}
\caption{\textbf{Camera pose accuracy on the DVLT benchmark.} \textbf{AUC@3} and \textbf{AUC@30} ($\uparrow$), mean $\pm$ std over three shifted evaluation seeds (\cref{s:benchmark}). \textbf{Mean}: relative change (\%) against the frozen model, computed per dataset and averaged equally over the four datasets; $^\dagger$: statistically better than the best baseline, with the 95\% stratified paired-bootstrap confidence interval excluding zero in the direction of improvement (\cref{s:ci}, \cref{tab:ci_supp}). Best of each block in \textbf{bold}. $^\star$: reimplemented following the original paper.}
\label{tab:dvlt_pose_std}
\resizebox{\textwidth}{!}{%
\begin{tabular}{l rr rr rr rr cc}
\toprule
\multirow{2}{*}[-4pt]{\textbf{Method}} & \multicolumn{2}{c}{\textbf{DTU} {\scriptsize(22)}} & \multicolumn{2}{c}{\textbf{ETH3D} {\scriptsize(13)}} & \multicolumn{2}{c}{\textbf{7-Scenes} {\scriptsize(18)}} & \multicolumn{2}{c}{\textbf{ScanNet++} {\scriptsize(50)}} & \multicolumn{2}{c}{\textbf{Mean} {\scriptsize(rel.\,\%)}} \\
\cmidrule(lr){2-3}\cmidrule(lr){4-5}\cmidrule(lr){6-7}\cmidrule(lr){8-9}\cmidrule(lr){10-11}
\multicolumn{1}{r}{\normalfont AUC} & @3$\uparrow$ & @30$\uparrow$ & @3$\uparrow$ & @30$\uparrow$ & @3$\uparrow$ & @30$\uparrow$ & @3$\uparrow$ & @30$\uparrow$ & @3$\uparrow$ & @30$\uparrow$ \\
\midrule
VGGT~\cite{vggt} & 95.6{\tiny$\pm$2.2} & 99.8{\tiny$\pm$0.1} & 35.3{\tiny$\pm$0.5} & 82.4{\tiny$\pm$0.4} & 11.5{\tiny$\pm$0.2} & 79.7{\tiny$\pm$0.1} & 16.0{\tiny$\pm$0.7} & 76.3{\tiny$\pm$0.1} & -- & -- \\
\quad + Test3R$^\star$~\cite{test3r} & 96.7{\tiny$\pm$1.6} & \textbf{99.9}{\tiny$\pm$0.1} & 36.6{\tiny$\pm$0.7} & 82.0{\tiny$\pm$0.8} & 11.5{\tiny$\pm$0.4} & 79.8{\tiny$\pm$0.1} & 15.8{\tiny$\pm$0.6} & 75.6{\tiny$\pm$0.1} & +0.9{\tiny$\pm$0.1} & $-$0.3{\tiny$\pm$0.2} \\
\quad + TCO~\cite{tco} & 88.8{\tiny$\pm$0.6} & 99.3{\tiny$\pm$0.1} & 30.7{\tiny$\pm$2.2} & 81.0{\tiny$\pm$0.8} & 11.5{\tiny$\pm$0.1} & 79.8{\tiny$\pm$0.0} & 17.3{\tiny$\pm$0.1} & 76.7{\tiny$\pm$0.0} & $-$2.8{\tiny$\pm$3.0} & $-$0.4{\tiny$\pm$0.3} \\
\quad + Self-Geometry$^\star$~\cite{selfgeo} & 78.2{\tiny$\pm$0.8} & 97.5{\tiny$\pm$0.1} & 22.4{\tiny$\pm$0.2} & 71.1{\tiny$\pm$1.3} & 7.3{\tiny$\pm$0.2} & 69.2{\tiny$\pm$0.3} & 10.6{\tiny$\pm$0.3} & 68.3{\tiny$\pm$0.7} & $-$31.1{\tiny$\pm$1.1} & $-$9.9{\tiny$\pm$0.2} \\
\rowcolor{oursrow}
\quad + Ours & \textbf{96.8}{\tiny$\pm$0.3} & \textbf{99.9}{\tiny$\pm$0.0} & \textbf{41.3}{\tiny$\pm$1.0} & \textbf{86.0}{\tiny$\pm$0.9} & \textbf{12.3}{\tiny$\pm$0.1} & \textbf{80.6}{\tiny$\pm$0.1} & \textbf{21.7}{\tiny$\pm$0.2} & \textbf{80.8}{\tiny$\pm$0.3} & \textbf{+15.5}$^\dagger${\tiny$\pm$2.8} & \textbf{+2.8}$^\dagger${\tiny$\pm$0.3} \\
\midrule
Depth Anything 3-Giant~\cite{da3} & 85.9{\tiny$\pm$6.0} & 99.0{\tiny$\pm$0.6} & 58.8{\tiny$\pm$0.6} & 81.6{\tiny$\pm$0.0} & 13.6{\tiny$\pm$0.1} & \textbf{81.3}{\tiny$\pm$0.1} & 21.9{\tiny$\pm$0.5} & 78.1{\tiny$\pm$0.4} & -- & -- \\
\quad + Test3R$^\star$~\cite{test3r} & 86.8{\tiny$\pm$2.7} & 99.1{\tiny$\pm$0.2} & 55.8{\tiny$\pm$1.3} & 80.7{\tiny$\pm$0.8} & 13.0{\tiny$\pm$0.2} & 80.0{\tiny$\pm$0.4} & 20.4{\tiny$\pm$1.1} & 74.0{\tiny$\pm$0.8} & $-$3.5{\tiny$\pm$2.4} & $-$1.9{\tiny$\pm$0.6} \\
\quad + TCO~\cite{tco} & 87.1{\tiny$\pm$4.9} & 99.1{\tiny$\pm$0.5} & 59.0{\tiny$\pm$0.7} & 81.6{\tiny$\pm$0.1} & \textbf{13.7}{\tiny$\pm$0.1} & \textbf{81.3}{\tiny$\pm$0.1} & 22.9{\tiny$\pm$0.5} & 78.5{\tiny$\pm$0.5} & +1.9{\tiny$\pm$0.3} & +0.2{\tiny$\pm$0.1} \\
\quad + Self-Geometry$^\star$~\cite{selfgeo} & 84.8{\tiny$\pm$4.6} & 98.9{\tiny$\pm$0.5} & 43.8{\tiny$\pm$2.4} & 78.9{\tiny$\pm$1.1} & 12.2{\tiny$\pm$0.3} & 78.0{\tiny$\pm$0.0} & 17.5{\tiny$\pm$1.2} & 75.1{\tiny$\pm$0.5} & $-$14.1{\tiny$\pm$1.3} & $-$2.8{\tiny$\pm$0.3} \\
\rowcolor{oursrow}
\quad + Ours & \textbf{91.1}{\tiny$\pm$0.8} & \textbf{99.6}{\tiny$\pm$0.1} & \textbf{63.1}{\tiny$\pm$0.5} & \textbf{84.5}{\tiny$\pm$0.1} & 13.6{\tiny$\pm$0.2} & \textbf{81.3}{\tiny$\pm$0.0} & \textbf{31.8}{\tiny$\pm$0.3} & \textbf{83.3}{\tiny$\pm$0.4} & \textbf{+14.9}$^\dagger${\tiny$\pm$2.3} & \textbf{+2.7}$^\dagger${\tiny$\pm$0.2} \\
\bottomrule
\end{tabular}}
\end{table*}

The three test-time adaptation baselines in \cref{tab:dvlt_main} differ in parameterizations and supervision signals. All baselines use the DVLT evaluation harness: each scene starts from released weights, adapts on the same selected frames without ground truth, uses the same benchmark metrics and scene aggregation, and starts the next scene afresh.

\paragraph{TCO~\cite{tco}.}
For VGGT, we run the official TCO VGGT-LoRA codebase with its published per-dataset hyperparameters: $50$ steps at learning rate $2\times10^{-4}$ with multi-view weight $1.0$ on DTU, and $40$ steps with weight $0.2$ at $10^{-3}$ on 7-Scenes and $5\times10^{-4}$ on ETH3D. Because ScanNet++ is absent from the TCO paper (its ScanNet script targets pose estimation with a ground-truth depth prior), ScanNet++ uses the indoor video setting (7-Scenes). Rank-4 LoRA ($\alpha=16$) adapts attention and MLP projections in cross-view blocks in \texttt{bf16} as released. For DA3-G, the adaptation backbone also uses \texttt{bf16} autocast. We port the LoRA placement to DA3-G's alternating cross-view blocks (blocks $13$--$39$ of $40$), keeping the per-image encoder frozen. Following TCO's official prior weights (pose $1$, translation $2$, intrinsics $0.01$) and Self-Geometry's ground-truth-free evaluation protocol~\cite{selfgeo}, the frozen model's own camera predictions serve as test-time priors.

\paragraph{Test3R~\cite{test3r}.}
Test3R optimizes visual prompts via triplet point-map consistency; since its released code targets DUSt3R~\cite{dust3r}, we faithfully port its encoder prompt placement. For VGGT, $32$ zero-initialized prompt tokens are injected into the first $12$ of the $24$ DINOv2~\cite{dinov2} per-image encoder blocks as specified in the paper; for DA3-G, they enter the $13$ per-image encoder blocks preceding cross-view attention. The triplet loss compares shared-view point maps via VGGT's point head or DA3-G's unprojected depth. Following~\cite{test3r}, each scene uses up to $165$ triplets, AdamW ($\beta=(0.9,0.95)$, zero weight decay, no gradient clipping), one update per two triplets, and the published per-dataset learning rates ($4\times10^{-5}$ on DTU; $10^{-5}$ on ETH3D, 7-Scenes, and ScanNet++), scaled by a per-model factor tuned on a disjoint split (\cref{s:base_tune}).

\paragraph{Self-Geometry~\cite{selfgeo}.}
Since Self-Geometry has no released code, we carefully reimplement it from the paper and its supplementary material. SuperPoint keypoints matched by LightGlue are extracted once per scene and filtered on the frozen prediction via epipolar residuals followed by multi-view residuals (each filter uses $3\times1.4826$ times the median nonnegative residual of the retained matches for that pair). Target and source views follow the paper's rotation binning (eight $15^\circ$ bins up to $120^\circ$ plus one beyond). Each step optimizes multi-view consistency, epipolar consistency, photometric, edge-aware smoothness, and depth-consistency losses via dynamic weight averaging, with thresholds for the multi-view and epipolar losses calibrated on filtered frozen predictions and the depth-consistency threshold recalculated per step. Before summing the gradients, we remove the component of the multi-view gradient parallel to the epipolar gradient. Rank-64 LoRA adapts query-key-value attention projections ($18.9\,\text{M}$ parameters on VGGT; $15.7\,\text{M}$ on DA3-G, matching the paper) for $50$ iterations at learning rate $5\times10^{-5}$ with warm-up and cosine decay, returning the post-update iterate minimizing the geometric mean of the two consistency terms on a fixed view set.

\paragraph{Disjoint Baseline Tuning.}
\label{s:base_tune}
For methods ported to new backbones (Test3R on both models and TCO on DA3-G), learning-rate scale factors are selected on a strictly disjoint tuning split (one training sequence from each 7-Scenes room) and held fixed across all four test datasets and all three evaluation seeds; TCO on VGGT retains its official settings, and Self-Geometry retains its published hyperparameters.

\section{Multi-Seed Stability and Significance}
\label{s:results}

We report variation across frame grids and student-mask seeds, and compare VGGT and DA3-G with their frozen checkpoints and the strongest adaptation baselines (\cref{tab:dvlt_pose_std,tab:dvlt_pmap_std,tab:methodseed,tab:ci_supp}).

\begin{table*}[!t]
\centering
\footnotesize
\setlength{\tabcolsep}{2.6pt}
\caption{\textbf{Point-map accuracy on the DVLT benchmark.} \textbf{Rel.~L2} ($\downarrow$, negative relative change is an improvement) and inlier ratio \textbf{IR} (\%, $\uparrow$), mean $\pm$ std over three evaluation seeds; notation as in \cref{tab:dvlt_pose_std}.}
\label{tab:dvlt_pmap_std}
\resizebox{\textwidth}{!}{%
\begin{tabular}{l rr rr rr rr cc}
\toprule
\multirow{2}{*}[-4pt]{\textbf{Method}} & \multicolumn{2}{c}{\textbf{DTU} {\scriptsize(22)}} & \multicolumn{2}{c}{\textbf{ETH3D} {\scriptsize(13)}} & \multicolumn{2}{c}{\textbf{7-Scenes} {\scriptsize(18)}} & \multicolumn{2}{c}{\textbf{ScanNet++} {\scriptsize(50)}} & \multicolumn{2}{c}{\textbf{Mean} {\scriptsize(rel.\,\%)}} \\
\cmidrule(lr){2-3}\cmidrule(lr){4-5}\cmidrule(lr){6-7}\cmidrule(lr){8-9}\cmidrule(lr){10-11}
 & Rel.~L2$\downarrow$ & IR$\uparrow$ & Rel.~L2$\downarrow$ & IR$\uparrow$ & Rel.~L2$\downarrow$ & IR$\uparrow$ & Rel.~L2$\downarrow$ & IR$\uparrow$ & Rel.~L2$\downarrow$ & IR$\uparrow$ \\
\midrule
VGGT~\cite{vggt} & 0.010{\tiny$\pm$0.000} & 95.5{\tiny$\pm$0.1} & 0.056{\tiny$\pm$0.002} & 57.6{\tiny$\pm$0.9} & 0.043{\tiny$\pm$0.001} & 70.3{\tiny$\pm$0.6} & 0.041{\tiny$\pm$0.002} & 62.8{\tiny$\pm$0.1} & -- & -- \\
\quad + Test3R$^\star$~\cite{test3r} & 0.010{\tiny$\pm$0.000} & 95.4{\tiny$\pm$0.1} & 0.055{\tiny$\pm$0.002} & 59.3{\tiny$\pm$1.1} & 0.041{\tiny$\pm$0.001} & 71.1{\tiny$\pm$0.6} & 0.043{\tiny$\pm$0.002} & 61.7{\tiny$\pm$0.2} & $-$0.0{\tiny$\pm$2.8} & +0.6{\tiny$\pm$0.8} \\
\quad + TCO~\cite{tco} & \textbf{0.008}{\tiny$\pm$0.000} & \textbf{97.4}{\tiny$\pm$0.0} & 0.054{\tiny$\pm$0.007} & 52.2{\tiny$\pm$4.6} & \textbf{0.037}{\tiny$\pm$0.001} & \textbf{74.4}{\tiny$\pm$0.3} & 0.037{\tiny$\pm$0.000} & 69.0{\tiny$\pm$0.2} & $-$10.4{\tiny$\pm$5.1} & +2.1{\tiny$\pm$2.4} \\
\quad + Self-Geometry$^\star$~\cite{selfgeo} & 0.010{\tiny$\pm$0.000} & 96.7{\tiny$\pm$0.2} & 0.103{\tiny$\pm$0.009} & 36.2{\tiny$\pm$2.0} & 0.068{\tiny$\pm$0.002} & 45.6{\tiny$\pm$0.3} & 0.061{\tiny$\pm$0.004} & 49.2{\tiny$\pm$0.7} & +48.4{\tiny$\pm$5.8} & $-$23.2{\tiny$\pm$1.2} \\
\rowcolor{oursrow}
\quad + Ours & 0.010{\tiny$\pm$0.000} & 95.6{\tiny$\pm$0.1} & \textbf{0.046}{\tiny$\pm$0.005} & \textbf{67.7}{\tiny$\pm$0.6} & 0.042{\tiny$\pm$0.001} & 71.6{\tiny$\pm$0.3} & \textbf{0.032}{\tiny$\pm$0.001} & \textbf{74.6}{\tiny$\pm$0.6} & \textbf{$-$10.8}{\tiny$\pm$3.5} & \textbf{+9.6}$^\dagger${\tiny$\pm$0.9} \\
\midrule
Depth Anything 3-Giant~\cite{da3} & 0.012{\tiny$\pm$0.001} & 96.8{\tiny$\pm$0.2} & 0.146{\tiny$\pm$0.001} & 63.2{\tiny$\pm$0.6} & 0.037{\tiny$\pm$0.000} & 70.8{\tiny$\pm$0.8} & 0.042{\tiny$\pm$0.000} & 58.4{\tiny$\pm$0.5} & -- & -- \\
\quad + Test3R$^\star$~\cite{test3r} & \textbf{0.010}{\tiny$\pm$0.001} & \textbf{97.1}{\tiny$\pm$0.1} & 0.129{\tiny$\pm$0.001} & 63.0{\tiny$\pm$0.4} & 0.041{\tiny$\pm$0.001} & 68.8{\tiny$\pm$0.4} & 0.045{\tiny$\pm$0.000} & 56.4{\tiny$\pm$0.6} & $-$2.3{\tiny$\pm$0.3} & $-$1.6{\tiny$\pm$0.1} \\
\quad + TCO~\cite{tco} & \textbf{0.010}{\tiny$\pm$0.000} & \textbf{97.1}{\tiny$\pm$0.0} & 0.144{\tiny$\pm$0.003} & 64.5{\tiny$\pm$0.6} & \textbf{0.036}{\tiny$\pm$0.000} & \textbf{73.2}{\tiny$\pm$0.7} & 0.040{\tiny$\pm$0.001} & 59.7{\tiny$\pm$0.5} & $-$7.4{\tiny$\pm$0.4} & +1.9{\tiny$\pm$0.1} \\
\quad + Self-Geometry$^\star$~\cite{selfgeo} & 0.012{\tiny$\pm$0.001} & 96.7{\tiny$\pm$0.4} & 0.153{\tiny$\pm$0.002} & 52.2{\tiny$\pm$2.7} & 0.044{\tiny$\pm$0.000} & 66.4{\tiny$\pm$0.9} & 0.044{\tiny$\pm$0.001} & 56.9{\tiny$\pm$1.2} & +5.3{\tiny$\pm$0.5} & $-$6.6{\tiny$\pm$1.2} \\
\rowcolor{oursrow}
\quad + Ours & 0.011{\tiny$\pm$0.000} & \textbf{97.1}{\tiny$\pm$0.0} & \textbf{0.116}{\tiny$\pm$0.003} & \textbf{67.1}{\tiny$\pm$0.8} & \textbf{0.036}{\tiny$\pm$0.000} & 72.0{\tiny$\pm$0.6} & \textbf{0.033}{\tiny$\pm$0.000} & \textbf{66.8}{\tiny$\pm$0.6} & \textbf{$-$13.0}$^\dagger${\tiny$\pm$1.4} & \textbf{+5.6}$^\dagger${\tiny$\pm$0.2} \\
\bottomrule
\end{tabular}}
\end{table*}

\subsection{Multi-Seed Accuracy and Standard Deviation}
\label{s:seedvar}

\paragraph{Consistent Gains Across Every Shifted Seed.}
\cref{tab:dvlt_pose_std,tab:dvlt_pmap_std} demonstrate that Free Geometry improves both camera trajectories and dense 3D geometry consistently across all three evaluation seeds (\cref{s:benchmark}), rather than benefiting a single favorable frame subset. On VGGT~\cite{vggt}, adaptation raises mean relative pose accuracy by $+15.5{\pm}2.8\%$ in AUC@3 and $+2.8{\pm}0.3\%$ in AUC@30 while lowering point-map Rel.~L2 by $-10.8{\pm}3.5\%$ and boosting inlier ratio (IR) by $+9.6{\pm}0.9\%$; on DA3-G~\cite{da3}, it delivers $+14.9{\pm}2.3\%$ AUC@3, $+2.7{\pm}0.2\%$ AUC@30, $-13.0{\pm}1.4\%$ Rel.~L2, and $+5.6{\pm}0.2\%$ IR. On ScanNet++ (50 scenes), Free Geometry lifts AUC@3 from $16.0$ to $21.7$ ($+5.7$) and IR from $62.8\%$ to $74.6\%$ ($+11.8$\,pts) on VGGT, and lifts AUC@3 from $21.9$ to $31.8$ ($+9.9$) and IR from $58.4\%$ to $66.8\%$ ($+8.4$\,pts) on DA3-G. Consequently, self-supervision from internal consensus reliably strengthens both models across diverse indoor and outdoor geometries.

\paragraph{Variation Across Frame Grids on DTU.}
On DTU, DA3-G improves from $85.9{\pm}6.0$ to $91.1{\pm}0.8$ AUC@3. The standard deviation across evaluation seeds falls by a factor of $7.5$ ($6.0/0.8$). For VGGT, the cross-seed AUC@3 standard deviation drops from ${\pm}2.2$ to ${\pm}0.3$ on DTU and from ${\pm}0.7$ to ${\pm}0.2$ on ScanNet++.

\subsection{Variation Over Method Seeds}
\label{s:methodseed}

\begin{table}[t]
\centering
\footnotesize
\setlength{\tabcolsep}{6.5pt}
\caption{\textbf{Variation over method seeds.} Equal-weight mean relative change (\%) against the frozen model across the four DVLT datasets (as in the Mean column of \cref{tab:dvlt_main}), reported as mean $\pm$ std over three method seeds that redraw all partial-view student masks at fixed evaluation seed~0 ($103$ scenes; for Rel.~L2 a negative change is an improvement).}
\label{tab:methodseed}
\begin{tabular}{@{}lccc@{}}
\toprule
Model & AUC@3$\uparrow$ & Rel.~L2$\downarrow$ & IR$\uparrow$ \\
\midrule
VGGT~\cite{vggt}  & $+12.6\pm0.3$ & $-11.5\pm2.3$ & $+9.4\pm0.4$ \\
DA3-G~\cite{da3} & $+13.4\pm0.3$ & $-10.6\pm1.3$ & $+6.4\pm0.8$ \\
\bottomrule
\end{tabular}
\end{table}

\paragraph{Insensitivity to Stochastic Partial-View Masking.}
Whereas the 16-pass consensus subsets use a fixed seed (\cref{s:consensus}), each of the $T=20$ adaptation steps randomly draws $B=2$ partial inputs (\cref{s:method}). To isolate this algorithmic mask-sampling variance from external frame-selection variance (\cref{s:seedvar}), we hold the input frames fixed at evaluation seed~0 (the default benchmark grid) and re-run Free Geometry across three independent method seeds that redraw the partial inputs of all $20$ steps. \cref{tab:methodseed} reports the equal-weight mean relative change against the frozen model across the four datasets, matching the Mean definition of \cref{tab:dvlt_main}. At evaluation seed~0, VGGT achieves $+12.6{\pm}0.3\%$ in AUC@3, $-11.5{\pm}2.3\%$ in Rel.~L2, and $+9.4{\pm}0.4\%$ in IR, while DA3-G achieves $+13.4{\pm}0.3\%$ in AUC@3, $-10.6{\pm}1.3\%$ in Rel.~L2, and $+6.4{\pm}0.8\%$ in IR. The mean changes have the same direction as those across evaluation grids in \cref{tab:dvlt_pose_std,tab:dvlt_pmap_std}. For both models and the three metrics in the table, the standard deviation over method seeds is less than a quarter (at most one-fifth) of the gain, and on camera pose AUC@3 it is just ${\pm}0.3\%$---nearly an order of magnitude smaller than the standard deviation across evaluation grids (${\pm}2.8\%$ and ${\pm}2.3\%$). 

\begin{table*}[!t]
\centering\footnotesize\setlength{\tabcolsep}{4pt}
\caption{\textbf{Confidence intervals} of the Mean columns of \cref{tab:dvlt_main}: The frozen rows report percentage changes relative to the frozen model. The baseline rows compare the two methods on that same scale, in percentage points. Brackets give the paired 95\% intervals (stratified, paired bootstrap over scenes within each dataset and over the three evaluation seeds, 10{,}000 resamples; for Rel.~L2 a negative change is an improvement). Best baseline per column: VGGT: Test3R (AUC@3), Test3R (AUC@30), TCO (Rel.~L2), TCO (IR); DA3-G: TCO (AUC@3), TCO (AUC@30), TCO (Rel.~L2), TCO (IR).}
\label{tab:ci_supp}
\begin{tabular}{@{}ll cccc@{}}
\toprule
Model & Comparison & AUC@3$\uparrow$ & AUC@30$\uparrow$ & Rel.~L2$\downarrow$ & IR$\uparrow$ \\
\midrule
VGGT~\cite{vggt} & Ours vs.\ frozen & +15.5 {\scriptsize[+11.4, +20.6]} & +2.8 {\scriptsize[+2.1, +3.7]} & $-$10.8 {\scriptsize[$-$15.0, $-$6.1]} & +9.6 {\scriptsize[+7.1, +12.8]} \\
 & Ours vs.\ best baseline & +14.7 {\scriptsize[+9.9, +20.3]} & +3.1 {\scriptsize[+2.2, +4.2]} & $-$0.4 {\scriptsize[$-$18.0, +10.5]} & +7.5 {\scriptsize[+1.9, +13.6]} \\
DA3-G~\cite{da3} & Ours vs.\ frozen & +14.9 {\scriptsize[+11.6, +18.7]} & +2.7 {\scriptsize[+1.7, +4.5]} & $-$13.0 {\scriptsize[$-$17.6, $-$8.1]} & +5.6 {\scriptsize[+4.1, +7.7]} \\
 & Ours vs.\ best baseline & +13.0 {\scriptsize[+9.9, +16.6]} & +2.5 {\scriptsize[+1.6, +4.3]} & $-$5.7 {\scriptsize[$-$10.3, $-$0.5]} & +3.7 {\scriptsize[+2.1, +5.7]} \\
\bottomrule
\end{tabular}
\end{table*}

\subsection{Paired-Bootstrap Confidence Intervals}
\label{s:ci}

\paragraph{Comparison With Frozen Checkpoints.}
To verify that these gains reflect systematic scene-wide improvements rather than isolated outliers, \cref{tab:ci_supp} reports 95\% confidence intervals for the Mean columns of \cref{tab:dvlt_main}. Each interval is computed via a stratified, paired bootstrap with $10{,}000$ resamples that draws scenes with replacement within each dataset and draws the three evaluation seeds with replacement, evaluating all methods on identical resamples before taking the $2.5$th and $97.5$th percentiles. Across VGGT and DA3-G, Free Geometry achieves statistically significant gains over the frozen model on \textbf{all 8 of 8} model--metric pairs, with wide positive margins on fine-grained pose ($[+11.4, +20.6]\%$ on VGGT AUC@3; $[+11.6, +18.7]\%$ on DA3-G AUC@3) and dense geometry ($[+7.1, +12.8]\%$ on VGGT IR; $[-17.6, -8.1]\%$ on DA3-G Rel.~L2).

\paragraph{Comparison With the Strongest Baseline.}
Comparing directly against the strongest test-time adaptation baseline per column (marked $^\dagger$ in \cref{tab:dvlt_main}), Free Geometry's 95\% improvement interval strictly excludes zero on \textbf{7 of 8} model--metric pairs. The intervals exclude zero for all four DA3-G metrics and for VGGT AUC@3, AUC@30, and IR. For VGGT Rel.~L2, the interval for the difference from TCO includes zero ($[-18.0,+10.5]$ percentage points).

\section{Cross-Architecture Generalization}
\label{s:generalize_supp}

We report per-dataset results and confidence intervals for VGGT-$\Omega$, $\pi^3$, and DVLT, followed by the five-model averages (\cref{tab:ci_other,tab:generalize_full,tab:generalize_full_pmap}).

\begin{table*}[!t]
\centering\footnotesize\setlength{\tabcolsep}{5pt}
\caption{\textbf{Confidence intervals for generalization to additional feed-forward models} (\cref{sec:generalize}, \cref{tab:generalize}): equal-weight mean relative change (\%) against the frozen model across the four DVLT datasets, with 95\% confidence intervals from a stratified, paired bootstrap over scenes within each dataset and over the three evaluation seeds ($10{,}000$ resamples; for Rel.~L2 a negative change is an improvement). Every 95\% confidence interval strictly excludes zero across all 12 model--metric pairs.}
\label{tab:ci_other}
\begin{tabular}{@{}ll cccc@{}}
\toprule
Model & Comparison & AUC@3$\uparrow$ & AUC@30$\uparrow$ & Rel.~L2$\downarrow$ & IR$\uparrow$ \\
\midrule
VGGT-$\Omega$~\cite{vggtomega} & Ours vs.\ frozen & +7.7 {\scriptsize[+5.1, +11.3]} & +1.0 {\scriptsize[+0.7, +1.3]} & $-$4.6 {\scriptsize[$-$6.1, $-$3.1]} & +2.2 {\scriptsize[+1.4, +3.2]} \\
$\pi^3$~\cite{pi3} & Ours vs.\ frozen & +1.4 {\scriptsize[+0.2, +3.3]} & +0.2 {\scriptsize[+0.03, +0.4]} & $-$2.6 {\scriptsize[$-$3.8, $-$1.5]} & +1.0 {\scriptsize[+0.4, +1.8]} \\
DVLT~\cite{dvlt} & Ours vs.\ frozen & +4.4 {\scriptsize[+2.6, +6.6]} & +0.5 {\scriptsize[+0.3, +0.7]} & $-$2.3 {\scriptsize[$-$3.8, $-$0.6]} & +1.3 {\scriptsize[+0.3, +2.6]} \\
\bottomrule
\end{tabular}
\end{table*}

\subsection{Confidence Intervals on Additional Models}
\label{s:ci_other}

\paragraph{Confidence Intervals Relative to the Frozen Models.}
Evaluated under the exact same $10{,}000$-resample stratified paired-bootstrap protocol used for VGGT and DA3-G (\cref{s:ci}), Free Geometry achieves statistically significant improvements on every mean metric of VGGT-$\Omega$, $\pi^3$, and DVLT (\cref{tab:ci_other}, substantiating \cref{sec:generalize} of the main paper). Across all $12$ model--metric pairs, every 95\% confidence interval strictly excludes zero: VGGT-$\Omega$ achieves $+7.7\%$ $[+5.1, +11.3]$ in AUC@3, $+1.0\%$ $[+0.7, +1.3]$ in AUC@30, $-4.6\%$ $[-6.1, -3.1]$ in Rel.~L2, and $+2.2\%$ $[+1.4, +3.2]$ in IR; $\pi^3$ achieves $+1.4\%$ $[+0.2, +3.3]$, $+0.2\%$ $[+0.03, +0.4]$, $-2.6\%$ $[-3.8, -1.5]$, and $+1.0\%$ $[+0.4, +1.8]$; and DVLT achieves $+4.4\%$ $[+2.6, +6.6]$, $+0.5\%$ $[+0.3, +0.7]$, $-2.3\%$ $[-3.8, -0.6]$, and $+1.3\%$ $[+0.3, +2.6]$. 

\paragraph{Why Diverse Architectural Regimes Consistently Benefit.}
Crucially, the statistical gains in \cref{tab:ci_other} hold across three fundamentally different modeling paradigms, demonstrating that our two free teacher sources are intrinsic to multi-view feed-forward inference rather than tied to one architecture. First, although frozen VGGT-$\Omega$ already starts from a much stronger baseline than VGGT ($52.1$ vs.\ $39.6$ mean AUC@3 in \cref{tab:generalize,tab:teacher}), single-pass predictions still fluctuate across reference frames and view subsets (\cref{fig:intro_compare}), enabling consensus supervision to secure wide confidence intervals of $[+5.1, +11.3]\%$ in AUC@3 and $[-6.1, -3.1]\%$ in Rel.~L2. Second, on reference-free, permutation-equivariant $\pi^3$---whose coarse pose is already near saturation ($90.1$ mean AUC@30)---near-full subset consensus paired with partial-view completion still yields strictly positive pose intervals ($[+0.2, +3.3]\%$ AUC@3; $[+0.03, +0.4]\%$ AUC@30) alongside decisive dense point-map refinement ($[-3.8, -1.5]\%$ Rel.~L2; $[+0.4, +1.8]\%$ IR). Third, on looping-transformer DVLT, connecting predicted dense ray maps to our gauge-invariant pose objective via differentiable Kabsch rotation fitting (\cref{s:protocol}) cleanly propagates multi-view trajectory and depth consistency into normalization layers, yielding $[+2.6, +6.6]\%$ AUC@3 and $[+0.3, +2.6]\%$ IR gains. Thus, Free Geometry transfers across stronger backbones, equivariant transformers, and implicit ray-map parameterizations with the shared optimization schedule.

\begin{table*}[!t]
\centering
\footnotesize
\setlength{\tabcolsep}{2.6pt}
\caption{\textbf{Camera pose accuracy} of the three additional feed-forward models per dataset (DVLT benchmark, 32 frames, fp16, expanding \cref{tab:generalize}): AUC@3 and AUC@30 ($\uparrow$), mean $\pm$ std over three evaluation seeds, using the shared protocol and model interfaces in \cref{s:protocol}. \textbf{Mean}: relative change (\%) against the frozen model, computed per dataset and averaged over the four datasets with equal weight (95\% confidence intervals in \cref{tab:ci_other}). Better of the two rows in \textbf{bold}.}
\label{tab:generalize_full}
\resizebox{\textwidth}{!}{%
\begin{tabular}{l rr rr rr rr cc}
\toprule
\multirow{2}{*}[-4pt]{\textbf{Method}} & \multicolumn{2}{c}{\textbf{DTU} {\scriptsize(22)}} & \multicolumn{2}{c}{\textbf{ETH3D} {\scriptsize(13)}} & \multicolumn{2}{c}{\textbf{7-Scenes} {\scriptsize(18)}} & \multicolumn{2}{c}{\textbf{ScanNet++} {\scriptsize(50)}} & \multicolumn{2}{c}{\textbf{Mean} {\scriptsize(rel.\,\%)}} \\
\cmidrule(lr){2-3}\cmidrule(lr){4-5}\cmidrule(lr){6-7}\cmidrule(lr){8-9}\cmidrule(lr){10-11}
 & AUC@3$\uparrow$ & AUC@30$\uparrow$ & AUC@3$\uparrow$ & AUC@30$\uparrow$ & AUC@3$\uparrow$ & AUC@30$\uparrow$ & AUC@3$\uparrow$ & AUC@30$\uparrow$ & AUC@3$\uparrow$ & AUC@30$\uparrow$ \\
\midrule
VGGT-$\Omega$~\cite{vggtomega} & 86.0{\tiny$\pm$1.1} & 99.1{\tiny$\pm$0.1} & 66.7{\tiny$\pm$0.6} & 95.2{\tiny$\pm$0.3} & 12.8{\tiny$\pm$0.4} & 81.0{\tiny$\pm$0.0} & 43.0{\tiny$\pm$0.9} & 88.9{\tiny$\pm$0.2} & -- & -- \\
\rowcolor{oursrow}
\quad + Ours & \textbf{86.7}{\tiny$\pm$0.6} & \textbf{99.2}{\tiny$\pm$0.1} & \textbf{72.6}{\tiny$\pm$0.6} & \textbf{96.5}{\tiny$\pm$0.1} & \textbf{13.4}{\tiny$\pm$0.4} & \textbf{81.2}{\tiny$\pm$0.0} & \textbf{50.1}{\tiny$\pm$1.3} & \textbf{90.9}{\tiny$\pm$0.2} & \textbf{+7.7}{\tiny$\pm$0.3} & \textbf{+1.0}{\tiny$\pm$0.1} \\
\midrule
$\pi^3$~\cite{pi3} & 69.4{\tiny$\pm$0.6} & 97.2{\tiny$\pm$0.1} & 42.3{\tiny$\pm$0.9} & 84.6{\tiny$\pm$0.4} & 11.4{\tiny$\pm$0.2} & 81.6{\tiny$\pm$0.1} & 75.5{\tiny$\pm$1.0} & 97.1{\tiny$\pm$0.2} & -- & -- \\
\rowcolor{oursrow}
\quad + Ours & \textbf{69.5}{\tiny$\pm$0.7} & \textbf{97.3}{\tiny$\pm$0.1} & \textbf{43.3}{\tiny$\pm$0.9} & \textbf{84.9}{\tiny$\pm$0.4} & \textbf{11.6}{\tiny$\pm$0.3} & \textbf{81.6}{\tiny$\pm$0.1} & \textbf{76.5}{\tiny$\pm$0.2} & \textbf{97.5}{\tiny$\pm$0.1} & \textbf{+1.4}{\tiny$\pm$0.5} & \textbf{+0.2}{\tiny$\pm$0.0} \\
\midrule
DVLT~\cite{dvlt} & 82.0{\tiny$\pm$1.5} & 98.7{\tiny$\pm$0.2} & 65.3{\tiny$\pm$0.9} & 95.1{\tiny$\pm$0.2} & 14.3{\tiny$\pm$0.2} & \textbf{82.1}{\tiny$\pm$0.1} & 77.5{\tiny$\pm$0.3} & 97.9{\tiny$\pm$0.0} & -- & -- \\
\rowcolor{oursrow}
\quad + Ours & \textbf{83.3}{\tiny$\pm$0.6} & \textbf{98.8}{\tiny$\pm$0.1} & \textbf{71.7}{\tiny$\pm$1.3} & \textbf{96.5}{\tiny$\pm$0.3} & \textbf{14.4}{\tiny$\pm$0.2} & 81.9{\tiny$\pm$0.2} & \textbf{81.9}{\tiny$\pm$0.1} & \textbf{98.4}{\tiny$\pm$0.0} & \textbf{+4.4}{\tiny$\pm$0.8} & \textbf{+0.5}{\tiny$\pm$0.1} \\
\bottomrule
\end{tabular}}
\end{table*}

\begin{table*}[!t]
\centering
\footnotesize
\setlength{\tabcolsep}{2.6pt}
\caption{\textbf{Point-map accuracy} of the three additional feed-forward models per dataset (DVLT benchmark, 32 frames, fp16, expanding \cref{tab:generalize}): Rel.~L2 ($\downarrow$) and inlier ratio IR (\%, $\uparrow$), mean $\pm$ std over three evaluation seeds, using the shared protocol and model interfaces in \cref{s:protocol}. \textbf{Mean}: relative change (\%) against the frozen model, computed per dataset and averaged over the four datasets with equal weight (for Rel.~L2 a negative change is an improvement; 95\% confidence intervals in \cref{tab:ci_other}). Better of the two rows in \textbf{bold}.}
\label{tab:generalize_full_pmap}
\resizebox{\textwidth}{!}{%
\begin{tabular}{l rr rr rr rr cc}
\toprule
\multirow{2}{*}[-4pt]{\textbf{Method}} & \multicolumn{2}{c}{\textbf{DTU} {\scriptsize(22)}} & \multicolumn{2}{c}{\textbf{ETH3D} {\scriptsize(13)}} & \multicolumn{2}{c}{\textbf{7-Scenes} {\scriptsize(18)}} & \multicolumn{2}{c}{\textbf{ScanNet++} {\scriptsize(50)}} & \multicolumn{2}{c}{\textbf{Mean} {\scriptsize(rel.\,\%)}} \\
\cmidrule(lr){2-3}\cmidrule(lr){4-5}\cmidrule(lr){6-7}\cmidrule(lr){8-9}\cmidrule(lr){10-11}
 & Rel.~L2$\downarrow$ & IR$\uparrow$ & Rel.~L2$\downarrow$ & IR$\uparrow$ & Rel.~L2$\downarrow$ & IR$\uparrow$ & Rel.~L2$\downarrow$ & IR$\uparrow$ & Rel.~L2$\downarrow$ & IR$\uparrow$ \\
\midrule
VGGT-$\Omega$~\cite{vggtomega} & 0.009{\tiny$\pm$0.000} & 97.2{\tiny$\pm$0.0} & 0.022{\tiny$\pm$0.000} & 83.1{\tiny$\pm$0.5} & 0.043{\tiny$\pm$0.001} & 71.8{\tiny$\pm$0.5} & 0.024{\tiny$\pm$0.001} & 83.1{\tiny$\pm$0.6} & -- & -- \\
\rowcolor{oursrow}
\quad + Ours & \textbf{0.009}{\tiny$\pm$0.000} & \textbf{97.2}{\tiny$\pm$0.0} & \textbf{0.019}{\tiny$\pm$0.000} & \textbf{86.0}{\tiny$\pm$0.1} & \textbf{0.042}{\tiny$\pm$0.001} & \textbf{72.8}{\tiny$\pm$0.1} & \textbf{0.022}{\tiny$\pm$0.001} & \textbf{86.3}{\tiny$\pm$0.5} & \textbf{$-$4.6}{\tiny$\pm$0.1} & \textbf{+2.2}{\tiny$\pm$0.2} \\
\midrule
$\pi^3$~\cite{pi3} & 0.009{\tiny$\pm$0.000} & 97.3{\tiny$\pm$0.0} & 0.034{\tiny$\pm$0.000} & 66.5{\tiny$\pm$0.2} & 0.032{\tiny$\pm$0.000} & 78.0{\tiny$\pm$0.3} & 0.015{\tiny$\pm$0.001} & 93.8{\tiny$\pm$0.4} & -- & -- \\
\rowcolor{oursrow}
\quad + Ours & \textbf{0.009}{\tiny$\pm$0.000} & \textbf{97.3}{\tiny$\pm$0.0} & \textbf{0.033}{\tiny$\pm$0.001} & \textbf{68.3}{\tiny$\pm$0.5} & \textbf{0.031}{\tiny$\pm$0.000} & \textbf{78.7}{\tiny$\pm$0.1} & \textbf{0.014}{\tiny$\pm$0.000} & \textbf{94.2}{\tiny$\pm$0.3} & \textbf{$-$2.6}{\tiny$\pm$0.3} & \textbf{+1.0}{\tiny$\pm$0.1} \\
\midrule
DVLT~\cite{dvlt} & 0.009{\tiny$\pm$0.000} & 97.0{\tiny$\pm$0.1} & 0.027{\tiny$\pm$0.000} & 76.7{\tiny$\pm$0.5} & 0.035{\tiny$\pm$0.001} & 74.6{\tiny$\pm$0.4} & 0.017{\tiny$\pm$0.000} & 92.1{\tiny$\pm$0.1} & -- & -- \\
\rowcolor{oursrow}
\quad + Ours & \textbf{0.009}{\tiny$\pm$0.000} & \textbf{97.0}{\tiny$\pm$0.1} & \textbf{0.025}{\tiny$\pm$0.000} & \textbf{80.0}{\tiny$\pm$0.1} & \textbf{0.035}{\tiny$\pm$0.000} & \textbf{75.1}{\tiny$\pm$0.3} & \textbf{0.017}{\tiny$\pm$0.000} & \textbf{92.1}{\tiny$\pm$0.1} & \textbf{$-$2.3}{\tiny$\pm$0.6} & \textbf{+1.3}{\tiny$\pm$0.2} \\
\bottomrule
\end{tabular}}
\end{table*}

\subsection{Per-Dataset Pose and Point-Map Breakdown}
\label{s:othermodels}

\paragraph{Adaptive Gains on Ambiguous Layouts With High Stability.}
\cref{tab:generalize_full,tab:generalize_full_pmap} expand \cref{tab:generalize} of the main paper into complete per-dataset camera pose and point-map metrics under the unified protocol (\cref{tab:impl}). The following comparisons use the frozen checkpoint for each model.
\begin{itemize}[leftmargin=1.2em, itemsep=2pt, topsep=2pt, parsep=0pt]
    \item \textbf{VGGT-$\Omega$~\cite{vggtomega}} improves or matches its frozen checkpoint on \emph{all 16 of 16 dataset--metric cells}. Through multi-order and subset consensus, it delivers major structural gains on wide-baseline ETH3D ($66.7\rightarrow 72.6$ AUC@3, $83.1\%\rightarrow 86.0\%$ IR, Rel.~L2 $0.022\rightarrow 0.019$) and 50-scene ScanNet++ ($43.0\rightarrow 50.1$ AUC@3, $83.1\%\rightarrow 86.3\%$ IR), while also lifting DTU pose ($86.0\rightarrow 86.7$ AUC@3, cutting std from ${\pm}1.1$ to ${\pm}0.6$) and 7-Scenes IR ($71.8\%\rightarrow 72.8\%$).
    \item \textbf{$\pi^3$~\cite{pi3}} likewise improves or matches the frozen model across \emph{all 16 of 16 dataset--metric cells}. Subset-consensus distillation raises ETH3D AUC@3 ($42.3\rightarrow 43.3$) and IR ($66.5\%\rightarrow 68.3\%$), lifts ScanNet++ AUC@3 ($75.5\rightarrow 76.5$, slashing cross-seed std $5\times$ from ${\pm}1.0$ to ${\pm}0.2$) and IR ($93.8\%\rightarrow 94.2\%$), and improves 7-Scenes Rel.~L2 and IR ($78.0\%\rightarrow 78.7\%$), confirming effective self-supervision without a reference frame.
    \item \textbf{DVLT~\cite{dvlt}} achieves substantial gains across complex indoor and outdoor benchmarks via differentiable ray-map supervision. Adaptation boosts ETH3D AUC@3 by $+6.4$ ($65.3\rightarrow 71.7$) and IR by $+3.3$\,pts ($76.7\%\rightarrow 80.0\%$), raises ScanNet++ AUC@3 by $+4.4$ ($77.5\rightarrow 81.9$, cutting std $3\times$ from ${\pm}0.3$ to ${\pm}0.1$), and lifts DTU AUC@3 by $+1.3$ ($82.0\rightarrow 83.3$, more than halving std ${\pm}1.5\rightarrow{\pm}0.6$) alongside $+0.5$\,pts IR on 7-Scenes ($74.6\%\rightarrow 75.1\%$).
\end{itemize}

\subsection{Statistical Synthesis Across All Five Models}
\label{s:five_model_synthesis}

\paragraph{20/20 Statistically Significant Model--Metric Gains.}
Taken together, \cref{tab:ci_supp} (VGGT and DA3-G) and \cref{tab:ci_other} (VGGT-$\Omega$, $\pi^3$, and DVLT) report the four mean metrics for each of the five models. Across all \textbf{20 of 20} model--metric pairs (five architectures $\times$ four pose and point-map metrics), the 95\% stratified paired-bootstrap confidence intervals of Free Geometry's improvements over the frozen model strictly exclude zero, with five-model average relative changes of \textbf{$+8.8\%$ in camera pose AUC@3}, \textbf{$+1.4\%$ in AUC@30}, \textbf{$-6.7\%$ in point-map Rel.~L2}, and \textbf{$+3.9\%$ in point-map inlier ratio (IR)}---directly substantiating the radar summary in \cref{fig:teaser} and headline claims of the main paper. One unified, label-free adaptation protocol thus consistently upgrades modern 3D foundation models across scales and architectures.

\section{Extended Qualitative Analysis}
\label{s:qual}
\begin{figure*}[!t]
\centering
\includegraphics[width=0.89\textwidth]{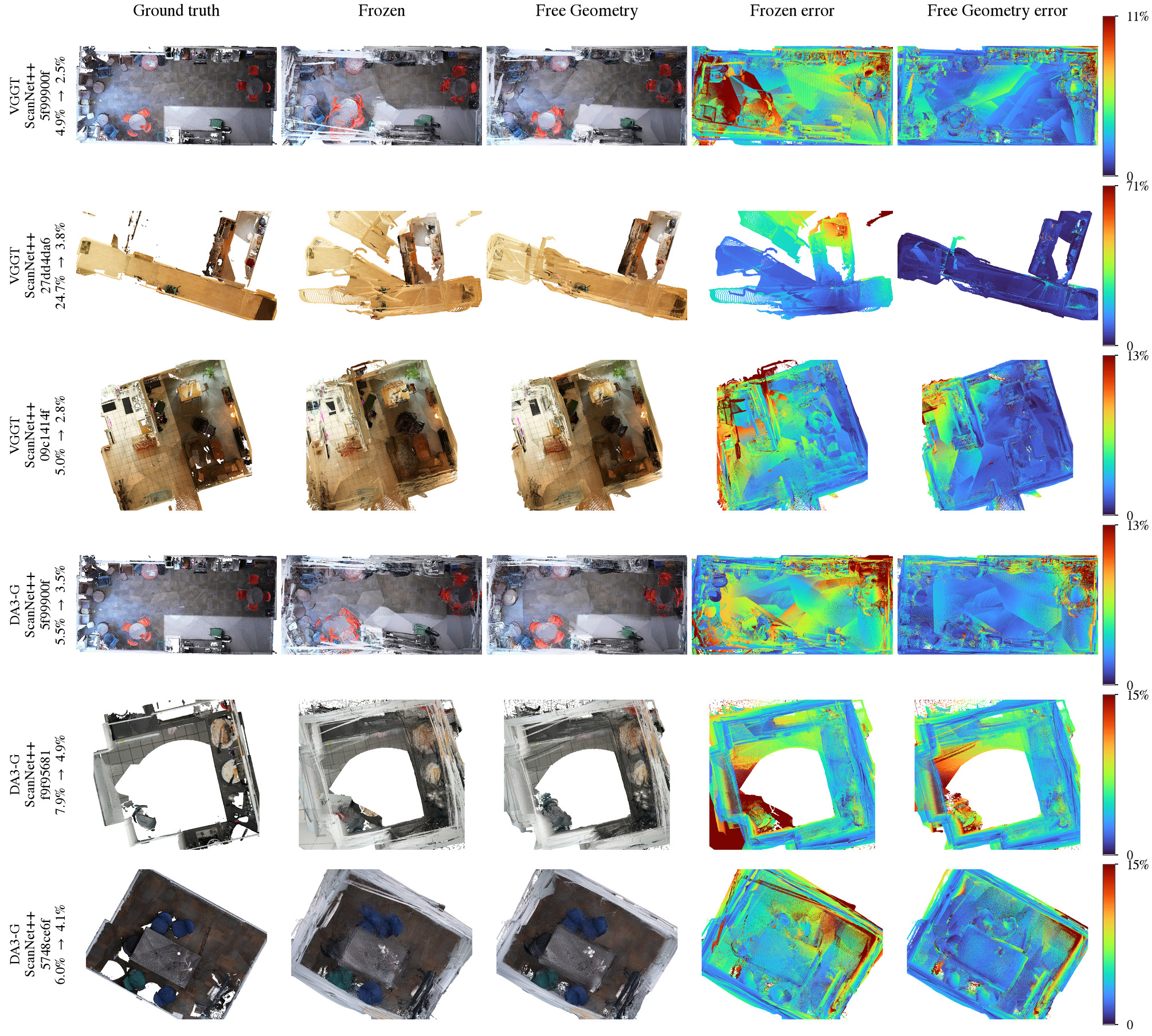}
\caption{\textbf{Point clouds of VGGT (top three rows) and DA3-G (bottom three rows)}: gravity-aligned top views of the fused point maps of all 32 frames, aligned to the ground truth as in the benchmark (points above the 85th height percentile, the ceiling, are removed). Columns: ground truth, frozen model and Free Geometry in colour, then the point error $e(p)$ of the frozen model and of Free Geometry, with a colour range per scene shared by both. The row label gives the mean point error. In the ScanNet++ lounge (first and fourth rows), the frozen models place the round table and the red chairs twice; Free Geometry merges the copies.}
\label{fig:s_qual_pc}
\end{figure*}

We show frozen and adapted predictions for the same scenes: point-map error maps for all five models and top-down point clouds for VGGT and DA3-G.

\subsection{Per-Pixel Error Maps on VGGT and DA3-G}
\label{s:qual_err_main}

\paragraph{Rescuing Global Misalignment and Local Drift on VGGT.}
\cref{fig:s_qual_err_vggt} shows per-pixel relative point-map errors $e(p)$ for the selected outdoor and indoor scenes (two frames per scene, shared colour range capped at the 95th percentile of frozen error). By aligning partial-view predictions to an outlier-trimmed 32-frame consensus, adaptation resolves severe cross-view drift: on ScanNet++ \texttt{27dd4da6}, where frozen VGGT suffers catastrophic global misalignment (scene mean $24.7\%$, individual frames $96.9\%$ and $77.3\%$), Free Geometry restores coherent geometry and slashes scene error to $3.8\%$ (frames $3.3\%$ and $7.3\%$). Substantial, scene-wide reductions hold across all other sequences as well---\texttt{6115eddb} drops from $5.8\%$ to $3.1\%$ (frames $30.7\%\rightarrow 6.6\%$ and $18.5\%\rightarrow 3.2\%$), \texttt{09c1414f} from $5.0\%$ to $2.8\%$, \texttt{5f99900f} from $4.9\%$ to $2.5\%$, and outdoor ETH3D \texttt{facade} and \texttt{delivery area} from $4.9\%\rightarrow 3.0\%$ and $4.2\%\rightarrow 2.6\%$---with lower scene-mean error in each example.

\paragraph{Point-Map Errors on DA3-G.}
\cref{fig:s_qual_err_da3} shows that the identical normalization-only updates eliminate severe depth and pose inconsistencies on the larger ViT-g DA3-G backbone. In ScanNet++ \texttt{f9f95681}, consensus self-supervision rescues a heavily drifted view from $92.5\%$ error down to $9.8\%$, lowering overall scene mean error from $7.9\%$ to $4.9\%$. Across indoor rooms \texttt{5f99900f} ($5.5\%\rightarrow 3.5\%$, frames $9.7\%\rightarrow 2.3\%$ and $8.5\%\rightarrow 1.8\%$), \texttt{5748ce6f} ($6.0\%\rightarrow 4.1\%$), \texttt{ac48a9b7} ($10.8\%\rightarrow 8.4\%$), \texttt{09c1414f} ($4.3\%\rightarrow 3.1\%$), and outdoor ETH3D \texttt{terrace} ($3.2\%\rightarrow 2.1\%$), the displayed scene means are lower after adaptation.

\subsection{Visual Gains on VGGT-$\Omega$, $\pi^3$, and DVLT}
\label{s:qual_err_other}

\paragraph{Structural Refinement Across Stronger and Distinct Paradigms.}
All 12 confidence intervals in \cref{tab:ci_other} exclude zero in the direction of improvement. The intervals for Rel.~L2 lie below zero, and those for AUC and IR lie above zero. \Cref{fig:s_qual_err_other} illustrates the corresponding changes in reconstruction. On VGGT-$\Omega$, adaptation tightens planar facades and indoor walls across ETH3D \texttt{facade} ($2.2\%\rightarrow 1.8\%$), ETH3D \texttt{relief} ($2.9\%\rightarrow 2.3\%$, frame $13.3\%\rightarrow 7.6\%$), and ScanNet++ \texttt{27dd4da6} ($2.1\%\rightarrow 1.6\%$, frames $7.9\%\rightarrow 1.7\%$ and $3.1\%\rightarrow 0.6\%$); on permutation-equivariant $\pi^3$, subset-consensus supervision removes doorway and wall misalignments in ScanNet++ \texttt{27dd4da6} ($3.2\%\rightarrow 1.4\%$, frames $9.0\%\rightarrow 1.8\%$ and $8.3\%\rightarrow 1.7\%$) and \texttt{f9f95681} ($1.5\%\rightarrow 0.7\%$); and on ray-map-based DVLT, ETH3D \texttt{delivery area} error drops from $2.5\%$ to $1.7\%$ (frames $4.2\%\rightarrow 2.2\%$ and $4.3\%\rightarrow 2.4\%$). Self-evolution thus sharpens fine-grained 3D geometry across every tested model family.

\subsection{Global 3D Point-Cloud Geometry Correction}
\label{s:qual_pc}

\paragraph{Eliminating Loop Drift, Room Folding, and Duplication.}
\cref{fig:s_qual_pc} presents gravity-aligned top-down views of fused 32-frame 3D point clouds alongside ground truth and per-point error maps for VGGT (top three rows) and DA3-G (bottom three rows), showing that Free Geometry repairs global topological failures caused by accumulated camera drift:
\begin{itemize}[leftmargin=1.2em, itemsep=1.5pt, topsep=2pt, parsep=0pt]
    \item \emph{Merging duplicated structures:} In the ScanNet++ lounge (\texttt{5f99900f}, rows 1 and 4), loop-closure drift causes both frozen VGGT ($4.9\%$) and frozen DA3-G ($5.5\%$) to hallucinate two separate copies of the central round table and red chairs; enforcing cross-frame relational pose and depth consistency aligns the loop and cleanly merges every duplicated object ($2.5\%$ and $3.5\%$).
    \item \emph{Unfolding severe global distortion:} In the elongated office corridor (\texttt{27dd4da6}, row 2), frozen VGGT ($24.7\%$) folds and fractures the room into overlapping segments; Free Geometry unfolds the entire trajectory to recover the true rectangular layout ($3.8\%$).
    \item \emph{Sharpening global room boundaries:} Across \texttt{09c1414f}, \texttt{f9f95681}, and \texttt{5748ce6f}, adaptation straightens orthogonal walls, removes ghosted double surfaces, and reduces point-cloud error in the illustrated room regions.
\end{itemize}

\section{Out-of-Domain and Degraded Scenes}
\label{s:robust}

We evaluate VGGT and SelfEvo on OmniGeo and report results on occluded, transparent-object and optically degraded scenes (\cref{tab:ood,tab:hard,tab:s_realx3d}).

\subsection{Out-of-Domain Evaluation: OmniGeo}
\label{s:robust_omni}

\paragraph{OmniGeo Evaluation.}
We follow the OmniGeo protocol of SelfEvo~\cite{selfevo} on OmniWorld-Game~\cite{omniworld}. \Cref{tab:ood} reports pairwise pose AUC at $5^\circ$ and $30^\circ$, and depth AbsRel and $\delta<1.25$ after per-sequence scale-and-shift alignment.

\paragraph{Zero-Shot Transfer and Complementarity With Offline Post-Training.}
We apply Free Geometry to both VGGT and SelfEvo on each OmniGeo test sequence (\cref{tab:ood}). First, Free Geometry adapts VGGT on the current test sequence without prior training on target-domain videos. Relative to frozen VGGT, adaptation raises AUC@5 from $45.2$ to $52.7$ ($+7.5$) and AUC@30 from $72.7$ to $80.0$ ($+7.3$) while cutting depth AbsRel from $0.180\rightarrow 0.156$ and lifting $\delta<1.25$ from $0.760\rightarrow 0.811$. Second, applying Free Geometry at test time to SelfEvo---which has already been post-trained offline on target-domain videos---further raises AUC@5 from $58.3\rightarrow 64.4$ ($+6.1$), AUC@30 from $87.3\rightarrow 91.6$ ($+4.3$), and $\delta<1.25$ from $0.867\rightarrow 0.875$ while lowering AbsRel from $0.124\rightarrow 0.118$. Instance-specific test-time self-evolution therefore captures scene-level geometric constraints that offline domain-level adaptation leaves unresolved.

\subsection{Challenging Scenes and RealX3D Breakdown}
\label{s:robust_realx3d}

\begin{table}[t]
\centering
\footnotesize
\setlength{\tabcolsep}{3.2pt}
\caption{\textbf{RealX3D per capture condition} (degraded images; VGGT frozen and with Free Geometry, lengths scaled with the number of views).}
\label{tab:s_realx3d}
\resizebox{\linewidth}{!}{%
\begin{tabular}{l l cc cc}
\toprule
 & & \multicolumn{2}{c}{\textbf{Pose AUC}} & \multicolumn{2}{c}{\textbf{Point map}} \\
\cmidrule(lr){3-4}\cmidrule(lr){5-6}
\textbf{Condition} & \textbf{Method} & @3$\uparrow$ & @30$\uparrow$ & Rel.~L2$\downarrow$ & IR$\uparrow$ \\
\midrule
Smoke (8) & VGGT & 82.6 & 98.5 & 0.0347 & 74.7 \\
\rowcolor{oursrow}
 & + Ours & \textbf{83.4} & \textbf{98.6} & \textbf{0.0329} & \textbf{76.9} \\
Low light (9) & VGGT & 84.0 & 98.8 & 0.0367 & 80.8 \\
\rowcolor{oursrow}
 & + Ours & \textbf{85.7} & \textbf{99.0} & \textbf{0.0365} & \textbf{81.0} \\
Reflection (8) & VGGT & \textbf{85.0} & 98.8 & 0.0406 & 78.4 \\
\rowcolor{oursrow}
 & + Ours & 84.8 & 98.8 & \textbf{0.0397} & \textbf{78.9} \\
Defocus (mild) (8) & VGGT & 87.7 & 99.1 & 0.0253 & 88.7 \\
\rowcolor{oursrow}
 & + Ours & \textbf{88.5} & \textbf{99.2} & \textbf{0.0251} & \textbf{88.8} \\
Defocus (strong) (8) & VGGT & 82.2 & 98.6 & \textbf{0.0293} & \textbf{86.3} \\
\rowcolor{oursrow}
 & + Ours & \textbf{84.1} & \textbf{98.8} & 0.0295 & 85.9 \\
Motion blur (mild) (8) & VGGT & \textbf{89.6} & 99.3 & 0.0213 & 90.9 \\
\rowcolor{oursrow}
 & + Ours & 89.1 & 99.3 & 0.0213 & 90.9 \\
Motion blur (strong) (8) & VGGT & 83.2 & 98.6 & \textbf{0.0274} & \textbf{87.5} \\
\rowcolor{oursrow}
 & + Ours & \textbf{84.8} & \textbf{98.8} & 0.0275 & 87.4 \\
\bottomrule
\end{tabular}}
\end{table}

\paragraph{Resolving Occlusion, Transparency, and Severe Optical Degradations.}
When visual cues break down in partial views due to occlusions, refractions, or degraded optics, unoccluded or cleaner perspectives across the $16$-pass consensus supply reliable geometric supervision for masked students (\cref{s:benchmark}). In \cref{tab:hard} of the main paper, this cross-view completion lifts VGGT on $30\%$-occluded ETH3D~\cite{mge} by $+4.1$ AUC@3 ($34.6\rightarrow 38.7$) and $+8.9$\,pts IR ($53.5\%\rightarrow 62.4\%$), and boosts pose AUC@3 on transparent B-TOGE tabletop objects~\cite{weibel2023challenges,btoge} by $+13.4$ ($65.4\rightarrow 78.8$) while improving object point maps. Decomposing the RealX3D~\cite{realx3d} benchmark across all seven physical capture degradations (\cref{tab:s_realx3d}) gives the following changes by condition:
\begin{itemize}[leftmargin=1.2em, itemsep=1.5pt, topsep=2pt, parsep=0pt]
    \item \emph{Atmospheric scattering and low illumination:} Under \textbf{Smoke}, consensus supervision improves all four metrics simultaneously ($82.6\rightarrow 83.4$ AUC@3, Rel.~L2 $0.0347\rightarrow 0.0329$, IR $74.7\%\rightarrow 76.9\%$); under \textbf{Low light}, it lifts AUC@3 by $+1.7$ ($84.0\rightarrow 85.7$) alongside gains in AUC@30, Rel.~L2, and IR.
    \item \emph{Optical blur and specular reflections:} Under severe blur where single-pass pose drifts---such as \textbf{Defocus (strong)} and \textbf{Motion blur (strong)}---Free Geometry raises pose AUC@3 by $+1.9$ ($82.2\rightarrow 84.1$) and $+1.6$ ($83.2\rightarrow 84.8$), while \textbf{Defocus (mild)} improves all four metrics, \textbf{Reflection} refines dense point maps (Rel.~L2 $0.0406\rightarrow 0.0397$, IR $78.4\%\rightarrow 78.9\%$), and under \textbf{Motion blur (mild)} both models have $90.9\%$ IR and $99.3$ AUC@30.
\end{itemize}
Averaged across all seven conditions, Free Geometry improves every metric on RealX3D (\cref{tab:hard}) without condition-specific tuning.

\section{Multi-Dataset and Multi-Model Ablations}
\label{s:ablation_full}

We compare teacher construction, student length, and loss terms on VGGT and DA3-G across four DVLT datasets. The full-backbone accuracy comparison uses VGGT (\cref{tab:abl_vggt_perds,tab:abl_da3_perds,tab:loss_perds}).

\begin{table*}[!t]
\centering\footnotesize
\setlength{\tabcolsep}{3.5pt}\renewcommand{\arraystretch}{0.92}
\caption{\textbf{Ablations of Free Geometry on VGGT across all four datasets} (mean over three evaluation seeds). Every row is the adapted model; the blocks group configurations by the aspect under study, extending \cref{tab:ablation4} from ETH3D to all four benchmarks. Default: full-view consensus, $M=16$, all three losses, normalization layers. Best of each block per column in \textbf{bold}.}
\label{tab:abl_vggt_perds}
\resizebox{\textwidth}{!}{%
\begin{tabular}{l ccc ccc ccc ccc}
\toprule
 & \multicolumn{3}{c}{DTU} & \multicolumn{3}{c}{ETH3D} & \multicolumn{3}{c}{7-Scenes} & \multicolumn{3}{c}{ScanNet++} \\
\cmidrule(lr){2-4} \cmidrule(lr){5-7} \cmidrule(lr){8-10} \cmidrule(lr){11-13}
\textbf{Variant} & AUC@3$\uparrow$ & Rel.~L2$\downarrow$ & IR$\uparrow$ & AUC@3$\uparrow$ & Rel.~L2$\downarrow$ & IR$\uparrow$ & AUC@3$\uparrow$ & Rel.~L2$\downarrow$ & IR$\uparrow$ & AUC@3$\uparrow$ & Rel.~L2$\downarrow$ & IR$\uparrow$ \\
\midrule
Frozen VGGT & 95.6 & 0.0099 & 95.5 & 35.3 & 0.0556 & 57.6 & 11.5 & 0.0427 & 70.3 & 16.0 & 0.0413 & 62.8 \\
\midrule
\multicolumn{13}{l}{\emph{(a) Teacher views}} \\
\quad 16 frames & \textbf{96.9} & 0.0098 & 95.6 & 38.7 & 0.0498 & 66.1 & \textbf{12.6} & \textbf{0.0412} & \textbf{72.0} & 20.7 & \textbf{0.0314} & 74.7 \\
\quad 24 frames & 96.8 & 0.0098 & \textbf{95.7} & 40.4 & 0.0478 & 65.9 & 12.5 & 0.0416 & 71.8 & 21.4 & \textbf{0.0314} & \textbf{75.0} \\
\rowcolor{oursrow}
\quad all frames (default) & 96.8 & \textbf{0.0097} & 95.6 & \textbf{41.3} & \textbf{0.0460} & \textbf{67.7} & 12.3 & 0.0419 & 71.6 & \textbf{21.7} & 0.0318 & 74.6 \\
\midrule
\multicolumn{13}{l}{\emph{(b) Teacher aggregation}} \\
\quad single pass & 95.5 & 0.0099 & \textbf{95.6} & 35.4 & 0.0573 & 58.2 & 11.7 & 0.0427 & 70.9 & 15.9 & 0.0405 & 62.3 \\
\rowcolor{oursrow}
\quad consensus (default) & \textbf{96.8} & \textbf{0.0097} & \textbf{95.6} & \textbf{41.3} & \textbf{0.0460} & \textbf{67.7} & \textbf{12.3} & \textbf{0.0419} & \textbf{71.6} & \textbf{21.7} & \textbf{0.0318} & \textbf{74.6} \\
\midrule
\multicolumn{13}{l}{\emph{(c) Student length $M$ (frames)}} \\
\quad $M=4$ & 95.8 & 0.0099 & \textbf{95.6} & 35.3 & 0.0477 & 61.2 & 11.8 & 0.0431 & 70.5 & 19.9 & 0.0363 & 69.3 \\
\quad $M=8$ & 96.7 & 0.0099 & \textbf{95.6} & 37.9 & 0.0507 & 65.8 & 12.1 & 0.0424 & 71.2 & 21.4 & 0.0335 & 73.4 \\
\rowcolor{oursrow}
\quad $M=16$ (default) & \textbf{96.8} & \textbf{0.0097} & \textbf{95.6} & 41.3 & \textbf{0.0460} & \textbf{67.7} & 12.3 & 0.0419 & \textbf{71.6} & 21.7 & \textbf{0.0318} & \textbf{74.6} \\
\quad $M=32$ (no masking) & 96.5 & 0.0098 & \textbf{95.6} & \textbf{42.5} & 0.0475 & 65.6 & 12.0 & \textbf{0.0417} & 71.4 & \textbf{22.0} & 0.0324 & 73.7 \\
\midrule
\multicolumn{13}{l}{\emph{(d) Loss terms}} \\
\quad w/o $\Lpose$ & 96.6 & 0.0098 & \textbf{95.7} & 38.2 & 0.0540 & 64.4 & 11.4 & 0.0420 & 71.2 & 18.7 & 0.0333 & 72.0 \\
\quad w/o $\mathcal{L}_{\text{rel}}$ & 96.7 & 0.0098 & \textbf{95.7} & \textbf{41.3} & 0.0470 & 67.0 & \textbf{12.3} & 0.0420 & 71.3 & 21.8 & 0.0324 & 73.7 \\
\quad w/o $\mathcal{L}_{\text{depth}}$ & \textbf{96.8} & 0.0098 & 95.6 & 38.7 & 0.0573 & 63.5 & \textbf{12.3} & \textbf{0.0419} & \textbf{71.9} & \textbf{22.0} & 0.0329 & 73.7 \\
\rowcolor{oursrow}
\quad all three (default) & \textbf{96.8} & \textbf{0.0097} & 95.6 & \textbf{41.3} & \textbf{0.0460} & \textbf{67.7} & \textbf{12.3} & \textbf{0.0419} & 71.6 & 21.7 & \textbf{0.0318} & \textbf{74.6} \\
\midrule
\multicolumn{13}{l}{\emph{(e) Trainable parameters}} \\
\quad all backbone weights & \textbf{96.8} & 0.0098 & \textbf{95.6} & \textbf{42.9} & \textbf{0.0398} & \textbf{70.3} & \textbf{12.4} & \textbf{0.0419} & \textbf{71.7} & 21.2 & 0.0320 & \textbf{75.3} \\
\rowcolor{oursrow}
\quad normalization layers (default) & \textbf{96.8} & \textbf{0.0097} & \textbf{95.6} & 41.3 & 0.0460 & 67.7 & 12.3 & 0.0419 & 71.6 & 21.7 & \textbf{0.0318} & 74.6 \\
\bottomrule
\end{tabular}}
\end{table*}

\subsection{Full Ablation Protocol}
\label{s:abl_setup}

\Cref{tab:abl_vggt_perds,tab:abl_da3_perds} compare the following configurations over all four DVLT datasets and three evaluation seeds (\cref{s:setup}):
\begin{itemize}[leftmargin=1.2em, itemsep=1.5pt, topsep=2pt, parsep=0pt]
    \item \emph{(a) Teacher views:} Consensus constructed from $16$ random subsets of $16$ or $24$ frames in random order (without the benchmark-order pass), versus our default full-view $32$-frame consensus (\cref{s:consensus}).
    \item \emph{(b) Teacher aggregation:} Single benchmark-order forward pass of the frozen model on all $32$ frames as teacher, versus our $16$-pass outlier-trimmed consensus.
    \item \emph{(c) Student length $M$:} Fixed per-step budget of $32$ frames partitioned into eight students at $M=4$, four at $M=8$, two at $M=16$ (default), or one unmasked full-sequence student at $M=32$.
    \item \emph{(d) Loss terms:} One objective term omitted from \cref{eq:combine}, with the remaining two unit-normalized gradients summed identically.
    \item \emph{(e) Trainable parameters:} Full-backbone fine-tuning in \texttt{fp32} with learning rate $10^{-6}$ selected from $\{10^{-6},10^{-5},10^{-4}\}$ on the disjoint 7-Scenes tuning split (\cref{s:base_tune}), versus normalization layers only. Following \cref{tab:ablation4}, full fine-tuning is ablated on VGGT across all four datasets, while \cref{tab:cost3} quantifies its $3{,}900\times$ parameter and $\sim 3\times$ memory overhead on DA3-G.
\end{itemize}

\subsection{Teacher View Coverage and Consensus}
\label{s:abl_teacher}

\paragraph{Single-Pass and Consensus Teachers (Block b).}
Blocks (b) of \cref{tab:abl_vggt_perds,tab:abl_da3_perds} compare the default consensus teacher with a single benchmark-order pass. Consistent with \cref{sec:ablation}, supervising with a single frozen pass yields virtually no pose gain over the frozen model on VGGT (e.g., $15.9$ vs.\ $16.0$ AUC@3 on ScanNet++; $35.4$ vs.\ $35.3$ on ETH3D) and produces only minor point-map shifts, whereas our $16$-pass consensus lifts ScanNet++ AUC@3 to $21.7$ and IR to $74.6\%$ and ETH3D AUC@3 to $41.3$ and IR to $67.7\%$. The identical pattern holds on DA3-G, where consensus strictly outperforms single-pass supervision on \emph{all 12 of 12} dataset--metric cells (e.g., ETH3D AUC@3 $57.5\rightarrow 63.1$; ScanNet++ AUC@3 $23.0\rightarrow 31.8$ and IR $59.8\%\rightarrow 66.8\%$).

\paragraph{Full-View Teacher Context Progressively Strengthens Multi-Room Geometry (Block a).}
Blocks (a) of \cref{tab:abl_vggt_perds,tab:abl_da3_perds} further show that expanding teacher context from $16$ to $24$ to all $32$ frames monotonically improves global camera pose on wide-baseline and multi-room scenes across both architectures. Because wider viewpoint coverage supplies long-range loop and baseline constraints, ETH3D AUC@3 climbs steadily $38.7\rightarrow 40.4\rightarrow 41.3$ on VGGT and $60.3\rightarrow 61.3\rightarrow 63.1$ on DA3-G, while ScanNet++ AUC@3 rises $20.7\rightarrow 21.4\rightarrow 21.7$ on VGGT and $30.7\rightarrow 31.6\rightarrow 31.8$ on DA3-G (where full-view supervision achieves the best overall performance on all $12/12$ DA3-G cells).

\begin{table*}[!t]
\centering\footnotesize
\setlength{\tabcolsep}{3.5pt}\renewcommand{\arraystretch}{0.92}
\caption{\textbf{Ablations of Free Geometry on DA3-G across all four datasets} (mean over three evaluation seeds). Every row is the adapted model; the blocks group configurations by the aspect under study, as in \cref{tab:ablation4}. Full-backbone fine-tuning is ablated on VGGT (\cref{tab:abl_vggt_perds}), with its $3{,}900\times$ parameter and $\sim 3\times$ memory overhead on DA3-G quantified in \cref{tab:cost3}. Default: full-view consensus, $M=16$, all three losses, normalization layers. Best of each block per column in \textbf{bold}.}
\label{tab:abl_da3_perds}
\resizebox{\textwidth}{!}{%
\begin{tabular}{l ccc ccc ccc ccc}
\toprule
 & \multicolumn{3}{c}{DTU} & \multicolumn{3}{c}{ETH3D} & \multicolumn{3}{c}{7-Scenes} & \multicolumn{3}{c}{ScanNet++} \\
\cmidrule(lr){2-4} \cmidrule(lr){5-7} \cmidrule(lr){8-10} \cmidrule(lr){11-13}
\textbf{Variant} & AUC@3$\uparrow$ & Rel.~L2$\downarrow$ & IR$\uparrow$ & AUC@3$\uparrow$ & Rel.~L2$\downarrow$ & IR$\uparrow$ & AUC@3$\uparrow$ & Rel.~L2$\downarrow$ & IR$\uparrow$ & AUC@3$\uparrow$ & Rel.~L2$\downarrow$ & IR$\uparrow$ \\
\midrule
Frozen DA3-G & 85.9 & 0.0122 & 96.8 & 58.8 & 0.1457 & 63.2 & 13.6 & 0.0370 & 70.8 & 21.9 & 0.0420 & 58.4 \\
\midrule
\multicolumn{13}{l}{\emph{(a) Teacher views}} \\
\quad 16 frames & 90.8 & 0.0113 & 97.0 & 60.3 & 0.1247 & 66.4 & 13.4 & 0.0369 & 71.7 & 30.7 & 0.0336 & 66.5 \\
\quad 24 frames & 91.0 & 0.0112 & 97.0 & 61.3 & 0.1198 & 66.8 & 13.5 & 0.0367 & 71.8 & 31.6 & 0.0337 & 66.6 \\
\rowcolor{oursrow}
\quad all frames (default) & \textbf{91.1} & \textbf{0.0110} & \textbf{97.1} & \textbf{63.1} & \textbf{0.1160} & \textbf{67.1} & \textbf{13.6} & \textbf{0.0365} & \textbf{72.0} & \textbf{31.8} & \textbf{0.0334} & \textbf{66.8} \\
\midrule
\multicolumn{13}{l}{\emph{(b) Teacher aggregation}} \\
\quad single pass & 85.8 & 0.0119 & 96.9 & 57.5 & 0.1447 & 63.9 & 13.5 & 0.0373 & 70.5 & 23.0 & 0.0408 & 59.8 \\
\rowcolor{oursrow}
\quad consensus (default) & \textbf{91.1} & \textbf{0.0110} & \textbf{97.1} & \textbf{63.1} & \textbf{0.1160} & \textbf{67.1} & \textbf{13.6} & \textbf{0.0365} & \textbf{72.0} & \textbf{31.8} & \textbf{0.0334} & \textbf{66.8} \\
\midrule
\multicolumn{13}{l}{\emph{(c) Student length $M$ (frames)}} \\
\quad $M=4$ & 86.6 & 0.0117 & 96.9 & 58.5 & 0.1289 & 66.2 & 13.2 & 0.0366 & 71.6 & 29.5 & 0.0363 & 64.4 \\
\quad $M=8$ & 89.5 & 0.0112 & 97.0 & 60.2 & 0.1385 & 67.0 & 13.4 & 0.0368 & 71.5 & 30.9 & 0.0346 & 65.9 \\
\rowcolor{oursrow}
\quad $M=16$ (default) & \textbf{91.1} & \textbf{0.0110} & \textbf{97.1} & \textbf{63.1} & \textbf{0.1160} & \textbf{67.1} & \textbf{13.6} & \textbf{0.0365} & \textbf{72.0} & \textbf{31.8} & \textbf{0.0334} & \textbf{66.8} \\
\quad $M=32$ (no masking) & 90.7 & 0.0113 & 97.0 & 62.4 & 0.1194 & 66.5 & 13.3 & 0.0370 & 71.7 & 31.7 & 0.0338 & 66.2 \\
\midrule
\multicolumn{13}{l}{\emph{(d) Loss terms}} \\
\quad w/o $\Lpose$ & 87.6 & 0.0113 & 97.0 & 61.1 & 0.1409 & 66.0 & 13.3 & 0.0369 & 71.5 & 28.5 & 0.0367 & 64.0 \\
\quad w/o $\mathcal{L}_{\text{rel}}$ & 90.9 & 0.0112 & 97.0 & 62.6 & 0.1282 & 66.6 & 13.4 & 0.0368 & 71.6 & 31.5 & 0.0339 & 66.3 \\
\quad w/o $\mathcal{L}_{\text{depth}}$ & 91.0 & 0.0117 & 97.0 & 62.7 & 0.1218 & 66.6 & 13.4 & 0.0367 & 71.6 & 31.6 & 0.0342 & 65.7 \\
\rowcolor{oursrow}
\quad all three (default) & \textbf{91.1} & \textbf{0.0110} & \textbf{97.1} & \textbf{63.1} & \textbf{0.1160} & \textbf{67.1} & \textbf{13.6} & \textbf{0.0365} & \textbf{72.0} & \textbf{31.8} & \textbf{0.0334} & \textbf{66.8} \\
\bottomrule
\end{tabular}}
\end{table*}

\subsection{Student Masking Granularity}
\label{s:abl_student}

\paragraph{Moderate Half-Sequence Masking ($M=16$) Across Datasets.}
Block (c) of \cref{tab:abl_vggt_perds,tab:abl_da3_perds} demonstrates across datasets and backbones that student view masking must balance optimization learnability against cross-view geometric completion. With $M=4$, ETH3D AUC@3 is $35.3$ on VGGT and $58.5$ on DA3-G, compared with frozen values of $35.3$ and $58.8$. ScanNet++ IR is lower with $M=4$ than with $M=16$ on both models ($69.3\%$ vs.\ $74.6\%$ on VGGT and $64.4\%$ vs.\ $66.8\%$ on DA3-G). Conversely, removing masking ($M=32$) retains pure ensemble distillation for cameras on VGGT but forfeits cross-view completion supervision from a longer teacher at higher per-student memory and compute---lowering VGGT point-map IR across ETH3D ($67.7\%\rightarrow 65.6\%$), ScanNet++ ($74.6\%\rightarrow 73.7\%$), and 7-Scenes ($71.6\%\rightarrow 71.4\%$), and trailing $M=16$ on \emph{all 12 of 12} cells on DA3-G. On VGGT, $M=16$ gives higher IR than $M=32$ on three datasets, while $M=32$ gives higher AUC@3 on ETH3D and ScanNet++.

\subsection{Geometric Loss Synergy and Depth Fidelity}
\label{s:abl_loss}

\begin{table}[t]
\centering\footnotesize\setlength{\tabcolsep}{3pt}
\caption{\textbf{Loss terms per dataset} (block (d) of the ablation), VGGT, mean over three evaluation seeds: pose AUC@3~$\uparrow$, point-map Rel.~L2~$\downarrow$ and IR~$\uparrow$, per-frame depth AbsRel (\%)~$\downarrow$. Best per dataset in \textbf{bold}.}
\label{tab:loss_perds}
\resizebox{\linewidth}{!}{%
\begin{tabular}{llcccc}
\toprule
Dataset & Variant & AUC@3$\uparrow$ & Rel.~L2$\downarrow$ & IR$\uparrow$ & Depth$\downarrow$ \\
\midrule
\multirow{4}{*}{DTU} & \cellcolor{oursrow}all three (default) & \cellcolor{oursrow}\textbf{96.8} & \cellcolor{oursrow}\textbf{0.0097} & \cellcolor{oursrow}95.6 & \cellcolor{oursrow}\textbf{0.91} \\
 & w/o $\Lpose$ & 96.6 & 0.0098 & \textbf{95.7} & \textbf{0.91} \\
 & w/o $\mathcal{L}_{\text{rel}}$ & 96.7 & 0.0098 & \textbf{95.7} & \textbf{0.91} \\
 & w/o $\mathcal{L}_{\text{depth}}$ & \textbf{96.8} & 0.0098 & 95.6 & \textbf{0.91} \\
\midrule
\multirow{4}{*}{ETH3D} & \cellcolor{oursrow}all three (default) & \cellcolor{oursrow}\textbf{41.3} & \cellcolor{oursrow}\textbf{0.0460} & \cellcolor{oursrow}\textbf{67.7} & \cellcolor{oursrow}\textbf{3.01} \\
 & w/o $\Lpose$ & 38.2 & 0.0540 & 64.4 & 3.33 \\
 & w/o $\mathcal{L}_{\text{rel}}$ & \textbf{41.3} & 0.0470 & 67.0 & 3.08 \\
 & w/o $\mathcal{L}_{\text{depth}}$ & 38.7 & 0.0573 & 63.5 & 3.41 \\
\midrule
\multirow{4}{*}{7-Scenes} & \cellcolor{oursrow}all three (default) & \cellcolor{oursrow}\textbf{12.3} & \cellcolor{oursrow}\textbf{0.0419} & \cellcolor{oursrow}71.6 & \cellcolor{oursrow}5.63 \\
 & w/o $\Lpose$ & 11.4 & 0.0420 & 71.2 & 5.64 \\
 & w/o $\mathcal{L}_{\text{rel}}$ & \textbf{12.3} & 0.0420 & 71.3 & 5.63 \\
 & w/o $\mathcal{L}_{\text{depth}}$ & \textbf{12.3} & \textbf{0.0419} & \textbf{71.9} & \textbf{5.61} \\
\midrule
\multirow{4}{*}{ScanNet++} & \cellcolor{oursrow}all three (default) & \cellcolor{oursrow}21.7 & \cellcolor{oursrow}\textbf{0.0318} & \cellcolor{oursrow}\textbf{74.6} & \cellcolor{oursrow}\textbf{5.73} \\
 & w/o $\Lpose$ & 18.7 & 0.0333 & 72.0 & 5.76 \\
 & w/o $\mathcal{L}_{\text{rel}}$ & 21.8 & 0.0324 & 73.7 & 5.79 \\
 & w/o $\mathcal{L}_{\text{depth}}$ & \textbf{22.0} & 0.0329 & 73.7 & 6.19 \\
\bottomrule
\end{tabular}}
\end{table}

\paragraph{Complementary Supervision Across Pose, Topology, and Dense Depth.}
Block (d) of \cref{tab:abl_vggt_perds,tab:abl_da3_perds}, reinforced by the per-frame depth evaluation in \cref{tab:loss_perds}, compares the full objective with each two-term variant:
\begin{itemize}[leftmargin=1.2em, itemsep=1.5pt, topsep=2pt, parsep=0pt]
    \item \emph{Pairwise relative pose ($\Lpose$):} Among the three loss ablations, removing $\Lpose$ gives the largest AUC@3 drop on ScanNet++ ($21.7\rightarrow 18.7$ on VGGT and $31.8\rightarrow 28.5$ on DA3-G) and ETH3D ($41.3\rightarrow 38.2$ and $63.1\rightarrow 61.1$, respectively).
    \item \emph{Global camera relations ($\mathcal{L}_{\text{rel}}$):} Omitting $\mathcal{L}_{\text{rel}}$ weakens global camera-centre topology and camera-to-depth scale coupling, consistently lowering point-map IR on ETH3D ($67.7\%\rightarrow 67.0\%$ on VGGT; $67.1\%\rightarrow 66.6\%$ on DA3-G) and ScanNet++ ($74.6\%\rightarrow 73.7\%$; $66.8\%\rightarrow 66.3\%$) while degrading DA3-G ETH3D Rel.~L2 from $0.1160$ to $0.1282$.
    \item \emph{Intra-frame depth ($\mathcal{L}_{\text{depth}}$):} Dropping $\mathcal{L}_{\text{depth}}$ degrades both per-frame depth fidelity and fused 3D point maps on complex scenes: on ETH3D (\cref{tab:loss_perds}), VGGT depth AbsRel worsens from $3.01\%$ to $3.41\%$ and AUC@3 drops from $41.3$ to $38.7$; on 50-scene ScanNet++, including $\mathcal{L}_{\text{depth}}$ cuts VGGT per-frame depth AbsRel from $6.19\%$ to $5.73\%$ while lifting IR from $73.7\%$ to $74.6\%$ (Rel.~L2 $0.0329\rightarrow 0.0318$), and on DA3-G (\cref{tab:abl_da3_perds}) it raises ScanNet++ IR from $65.7\%$ to $66.8\%$ and improves DTU and ETH3D Rel.~L2.
\end{itemize}
On VGGT, the depth term reduces depth AbsRel on ETH3D and ScanNet++. On 7-Scenes, AbsRel is $5.63\%$ with all three terms and $5.61\%$ without the depth term.

\subsection{Normalization Layers vs.\ Full Fine-Tuning}
\label{s:abl_params}

\paragraph{Capturing Full-Model Adaptation Gains at a Tiny Fraction of the Cost.}
Block (e) of \cref{tab:abl_vggt_perds} reports VGGT accuracy. Full fine-tuning gives higher AUC@3 on ETH3D ($42.9$ vs.\ $41.3$), while normalization-only tuning gives higher AUC@3 on ScanNet++ ($21.7$ vs.\ $21.2$). On DTU the two settings tie on AUC@3 and IR. The separate DA3-G cost measurement in \cref{tab:cost3} gives $17.4\,\text{GB}$ peak memory for normalization-only tuning and $51.3\,\text{GB}$ for full fine-tuning.

\end{document}